\documentclass[11pt]{article}
\pdfoutput=1

\usepackage[preprint]{acl}

\usepackage{times}
\usepackage{latexsym}

\usepackage[T1]{fontenc}

\usepackage[utf8]{inputenc}

\usepackage{microtype}

\usepackage{inconsolata}
 
\usepackage{graphicx}
\usepackage{algorithm}
\usepackage{algpseudocode}
\usepackage{booktabs,chemformula}
\usepackage{amsfonts}
\usepackage[subrefformat=parens]{subcaption}
\usepackage{adjustbox}
\usepackage{cleveref}
\usepackage{makecell}
\usepackage{xcolor}
\usepackage{colortbl}
\usepackage{multirow}
\usepackage{mathabx}
\definecolor{bluee}{cmyk}{0.60, 0.19, 0.0, 0.08}
\definecolor{redd}{cmyk}{0.19, 0.60, 0.0, 0.08}
\definecolor{greene}{cmyk}{0.50, 0.0, 0.50, 0.15}
\usepackage{xcolor, tabularx, booktabs}
\usepackage{tcolorbox}
\tcbuselibrary{skins}
\usepackage{pifont}
\newcommand{\cmark}{\ding{51}}
\newcommand{\xmark}{\ding{55}}

\newtcolorbox{findingbox}[1]{colback=greene!8,colframe=greene!55,arc=1mm,boxrule=0.3pt,left=4pt,right=4pt,top=1pt,bottom=1pt,enhanced,fonttitle=\bfseries,title={#1},coltitle=black,colbacktitle=greene!8,attach title to upper={\ },bottomtitle=0pt,toptitle=0pt,before skip=2pt,after skip=2pt}
\newcommand{\finding}[2]{\begin{findingbox}{\textit{Finding #1:}}#2\end{findingbox}}

\newcolumntype{u}{>{\columncolor{bluee!8}}c}
\newcolumntype{v}{>{\columncolor{bluee!24}}c}
\newcolumntype{d}{>{\columncolor{redd!8}}c}

\usepackage{tikz}
\usetikzlibrary{calc}
\usetikzlibrary{arrows.meta}
\usepackage{listings}
\usepackage{upquote}

\title{KCSAT-ML: Probing Reasoning Models with Nationwide-Cohort Human Difficulty}

\author{Sanghee Park$^\ast$\\
  NAVER Cloud AI\\
  KAIST AI\\
  \texttt{parksangheeeee@gmail.com} \\\And
  Geewook Kim$^\ast$$^\dagger$\\
  NAVER Cloud AI\\
  KAIST AI\\
  \texttt{gwkim.rsrch@gmail.com} \\\And
  Kee-Eung Kim$^\dagger$\\
  KAIST AI\\
  \texttt{kekim@kaist.ac.kr} \\
  }

\begin{document}
\maketitle

\begingroup
\renewcommand\thefootnote{}%
\footnotetext{$^{\ast}$ Sanghee Park and Geewook Kim contributed equally to this work and share first authorship.}%
\footnotetext{$^{\dagger}$ Corresponding authors.}%
\endgroup

\begin{abstract}
Math reasoning benchmarks have proliferated, yet most lack a per-item difficulty signal grounded in actual human performance.
We introduce \textbf{KCSAT-ML}, a decade (2014--2025) of Korean College Scholastic Ability Test (KCSAT; \textit{Suneung}) mathematics: 664 problems with a 339-item core set carrying official per-item error rates from nationwide cohorts of hundreds of thousands of examinees.
We pair the benchmark with \emph{Difficulty-aligned Reasoning Gain} (DRG): a score-orthogonal metric that asks whether a model's mistakes concentrate on the items humans found hard, or on items humans found easy.
Together they expose, across a wide range of VLMs (and LLMs via OCR), three patterns: (i) low-budget accuracy collapses on the high-human-error tail at every model size; (ii) test-time scaling (TTS) raises token use roughly linearly with cohort error rate, while accuracy gains follow a non-monotonic curve; (iii) within a single family, TTS flips between \emph{anti-scaling} on the hardest items and \emph{overthinking} on easier ones --- two faces of the same alignment failure.
On DRG, models with near-identical accuracy can sit at near-opposite values: one model gets wrong what humans also find hard, while another solves the hardest items yet fails on items humans find easy --- a contrast that aggregate accuracy hides. Our code and dataset builder is fully open-sourced at \url{https://github.com/naver-ai/KCSAT-ML}.
\end{abstract}

\begin{figure}[t!]
  \centering
  \includegraphics[width=\linewidth]{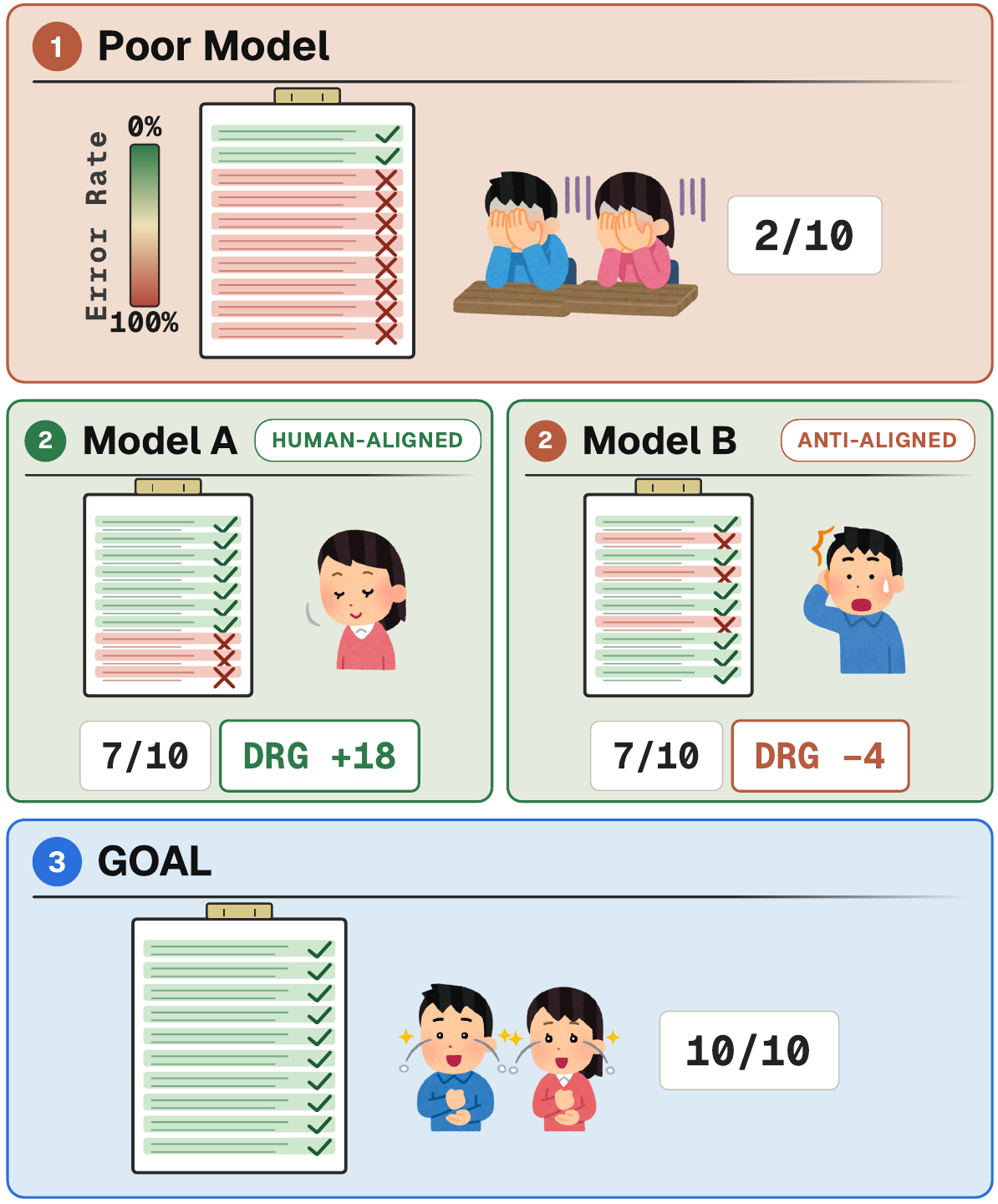}
    \caption{\textbf{Same score, opposite reasoning.} Two models both solve 7/10, but their mistakes land on opposite ends of the human-difficulty axis: Model A fails where humans also fail (DRG $+18$), Model B fails on items humans find easy (DRG $-4$). Accuracy alone cannot tell them apart; DRG (Sec.~\ref{sec:drg}) can.}
\label{fig:teaser}
\end{figure}

\section{Introduction}

Mathematical reasoning is now a central axis along which language and vision-language models are evaluated, and benchmarks across textual~\cite{hendrycks2021math}, multimodal~\cite{lu2023mathvista,qiao2025wemath}, and olympiad-level~\cite{he-etal-2024-olympiadbench,gao2024omnimath} settings have multiplied accordingly.
A benchmark, however, only resolves what its difficulty signal can resolve.
Most existing math benchmarks report accuracy aggregated across items without any per-item notion of how hard each problem actually is, and the few that do attach difficulty labels rely on heuristic ordinal annotations (e.g., the 1--5 levels in MATH~\cite{hendrycks2021math}) assigned by the authors themselves.
Without an item-level signal grounded in actual human performance, ``difficulty'' collapses into whatever the model finds hard.
This conflates problem difficulty with model-specific weaknesses, and leaves the field without a measurement-based axis for comparing model behaviour.

KoNET~\citep{park-kim-2025-evaluating} takes an important step in this direction by attaching official examinee statistics to items from the Korean College Scholastic Ability Test (KCSAT; \textit{Suneung}), Korea's annual high-stakes college-entrance examination.
Its mathematics subset, however, contains only 21 items with human error rates --- too few to characterise the high-difficulty tail or support fine-grained per-item analysis.
Moreover, the human signal is used primarily as additional reporting metadata, rather than as a tool for analysing how model behaviour varies with difficulty.
The field therefore lacks both a sufficiently large mathematics benchmark with item-level human statistics and tools to turn such statistics into an analysis axis.

We introduce \textbf{KCSAT-ML}, a decade (2014--2025) of KCSAT mathematics: 664 problems in total, with a 339-item core set carrying official per-item error rates measured from nationwide cohorts of hundreds of thousands of examinees.
Unlike author-assigned ordinal tiers, these rates are measured from examinee performance under standardised conditions, supplying a continuous, behaviourally grounded difficulty signal at a scale unavailable in prior math benchmarks.

This signal makes it possible to ask a question more demanding than aggregate accuracy: whether the items a model gets wrong are also the items humans find hard.
A model that solves an olympiad-level problem but stumbles on a textbook-level one is high-scoring yet unlike a human reasoner --- and arguably less trustworthy.
To make this notion measurable, we introduce \emph{Difficulty-aligned Reasoning Gain} (DRG): a score-orthogonal metric that captures whether additional inference-time compute aligns a model's failure pattern with the human difficulty axis, rather than merely clearing easy residual errors. DRG records \emph{where} the gains from added compute fall along that axis; it is not a measure of the quality of a model's internal reasoning.
We use this framework on KCSAT-ML to probe a question central to modern reasoning models: \emph{when} the extra compute spent by techniques such as Chain-of-Thought prompting~\cite{wei2022chain,wang2022selfconsistency} and \textit{test-time scaling} (TTS)~\cite{snell2024scaling,bi2024forestofthought,muennighoff-etal-2025-s1,deepseek2025r1} actually helps, and how its cost scales with the difficulty humans observe.

The pattern is sharp.
Low-budget accuracy collapses on the high-human-error tail at every model size, and TTS only partially recovers it: token use grows with cohort error rate, but accuracy gains follow a non-monotonic curve.
Within a single family, the same TTS turn-on yields opposite effects at the two extremes.
On the hardest items, \emph{anti-scaling} appears: a smaller w.\,TTS variant beats a larger wo.\,TTS sibling.
On easier items, \emph{overthinking} appears: extra reasoning derails an already-correct answer.
Both are alignment failures --- the model's competence and the human difficulty axis come apart, in opposite directions.
DRG turns this into a model-level diagnostic: across our 22 model families, pairs with near-identical accuracy can sit at near-opposite DRG values, showing that final scores can hide near-opposite failure distributions along the human difficulty axis.

Our contributions are as follows:
\begin{itemize}
    \item \textbf{KCSAT-ML, a human-grounded difficulty instrument.} We release KCSAT-ML (2014--2025; 664 items, 339 with official cohort error rates from hundreds of thousands of examinees) and show that the human signal surfaces model behaviour that author-assigned ordinal labels miss.
    \item \textbf{DRG: a metric for difficulty-aligned gains.} \emph{Difficulty-aligned Reasoning Gain} (DRG) is a score-orthogonal summary of whether additional inference-time compute aligns a model's failure pattern with the human difficulty axis, rather than merely clearing easy residual errors. On DRG, models with near-identical accuracy can sit far apart, and that placement is largely fixed at the very first TTS turn-on.
    \item \textbf{A difficulty-conditioned view of TTS.} Using DRG and the cohort signal, we map TTS behaviour across the difficulty range: a difficulty-dependent collapse--recovery, near-linear token cost in cohort error rate, and two asymmetric failure modes at the extremes (\emph{anti-scaling} on Hard, \emph{overthinking} on Easy).
\end{itemize}

\section{Related Work}
\label{sec:related_work}

\paragraph{Mathematical reasoning benchmarks.}
Existing math reasoning benchmarks span textual~\citep{hendrycks2021math}, multimodal~\citep{lu2023mathvista, qiao2025wemath}, competition~\citep{he-etal-2024-olympiadbench, gao2024omnimath}, and research-level mathematics.
Soohak~\citep{son2026soohak}, a 439-problem benchmark authored by 64 mathematicians, pushes evaluation past olympiad difficulty now that frontier models have reached IMO gold.
Most of these benchmarks rely on author-assigned ordinal tiers, or provide no per-item difficulty signal at all.
KoNET~\citep{park-kim-2025-evaluating} is closest in spirit to our work: it attaches nationwide-cohort statistics to Korean national exam items, but its KCSAT mathematics subset contains only 21 items with human error rates, and the signal is reported as metadata rather than used as an analysis axis.
KCSAT-ML extends this direction along two complementary axes --- a 339-item core set with cohort error rates from hundreds of thousands of examinees, and DRG, which turns the signal into a probe of where model failures and TTS gains concentrate along the human-difficulty axis, and at what cost.
Beyond a single label per item, each core item carries two independent signals: official point values (3 or 4) set by the exam designers, and item-level cohort error rates.

\paragraph{Human difficulty and psychometrics.}
Grounding item difficulty in examinee performance is the premise of educational measurement.
In Classical Test Theory (CTT), an item's difficulty is summarised by the proportion of examinees who answer it correctly---the \emph{item difficulty} (or \emph{facility}) index~\citep{lord1968statistical}---so the nationwide cohort error rate we build on is precisely the complement of this index, a standard psychometric quantity rather than a newly invented construct.
Item Response Theory (IRT) instead jointly estimates latent examinee ability together with per-item difficulty and discrimination from individual response patterns~\citep{embretson2000irt}, and has been used to make evaluation more informative: building evaluation scales for NLP~\citep{lalor-etal-2016-building}, estimating item parameters from model rather than human responses~\citep{lalor-etal-2019-learning}, analysing leaderboard reliability and example informativeness~\citep{rodriguez-etal-2021-evaluation}, selecting compact yet informative benchmark subsets~\citep{maiapolo2024tinybenchmarks}, and characterising classifiers at the instance level~\citep{martinezplumed2019irt}; predicting difficulty from item text is surveyed by \citet{benedetto2023survey}.
We deliberately adopt the CTT index for two reasons.
First, IRT calibration requires the full person-by-item response matrix, whereas KICE releases only aggregate per-item statistics, so the CTT index is the strongest human signal available for this exam.
Second, and more fundamentally, IRT difficulty in these settings is estimated \emph{from the responses of the systems under evaluation}, making it model-dependent and computed post hoc; the cohort error rate instead exists \emph{a priori}---measured on a human population before any model is run---which is what lets it serve as an external reference axis for model behaviour (Sec.~\ref{sec:diff_signal}).

\paragraph{Test-time scaling and inverse scaling.}
Test-time scaling extends inference compute via chain-of-thought~\citep{wei2022chain}, self-consistency~\citep{wang2022selfconsistency}, reasoning tokens~\citep{snell2024scaling, muennighoff-etal-2025-s1}, process-level verification~\citep{lightman2024verify}, and adaptive routing~\citep{liao2025rewardguided, zhang-etal-2025-adaptthink}.
Recent work has begun to document failure modes within this paradigm, including overthinking on simple problems in o1-like models~\citep{chen2024donot}.
Sec.~\ref{sec:two_faces} situates such failures along the human-difficulty axis: anti-scaling on the hardest items and overthinking on the easiest are two faces of the same alignment failure, jointly summarised by DRG.
The key distinction is the axis of failure: the inverse-scaling literature~\citep{mckenzie2023inverse} treats parameter count as the sole failure-inducing axis, whereas anti-scaling is a two-axis effect: at the high-difficulty tail, a smaller model with inference-time compute can reverse the parameter-count gap against a larger low-budget sibling.

\section{KCSAT-ML and Experimental Setup}
\label{sec:benchmark_construction}

\subsection{KCSAT-ML}

\textbf{KCSAT-ML} (KCSAT Mathematics Longitudinal) compiles a decade of Korean College Scholastic Ability Test (KCSAT) mathematics together with item-level error rates measured from hundreds of thousands of examinees. KCSAT items typically mix symbolic expressions with diagrams and graphs, so the benchmark is multimodal by default.
KCSAT-ML extends the KCSAT subset of KoNET~\citep{park-kim-2025-evaluating}, which contains 21 math items with human statistics, to 664 problems spanning 2014--2025, with official per-item error rates for a 339-item core subset.
KCSAT mathematics has been administered under several variants (e.g., Type-A and Type-B tracks before 2022, and a common-plus-elective structure from 2022 onward); we aggregate all publicly released variants within this period.
All evaluations and difficulty-conditioned analyses in this paper use the 339-item core set. The remaining items, generally easier problems for which item-level statistics are not publicly released, provide longitudinal coverage but are not used in our main experiments.

\paragraph{Construction.}
We collect official PDF booklets released by the Korea Institute of Curriculum and Evaluation (KICE)\footnote{\url{https://www.kice.re.kr}} and parse them into item-level instances. Each item is preserved as a single image of the original exam layout, matched with structured metadata (year, form, question number, point value, answer type) and, where available, official human statistics. This image-first format keeps the model input close to the setting in which cohort error rates were measured: VLMs are evaluated directly on the image, while OCR\footnote{\url{https://github.com/deepseek-ai/DeepSeek-OCR}} enables supplementary text-only LLM evaluation. We release annotations and scripts that reconstruct item images from the public KICE PDFs (details in Appendix~\ref{sec:dataset_construction}).

\paragraph{Human statistics.}
Each annual KCSAT yields item-level error rates from the nationwide cohort of roughly half a million examinees, providing an external, temporally consistent, continuous human-grounded difficulty signal. The official point values (3 or 4 in our core set) provide a complementary heuristic tier; Sec.~\ref{sec:diff_signal} quantifies the gap between the two signals.

\begin{table}[t]
\centering
\begin{adjustbox}{width=\linewidth}
\begin{tabular}{lccc}
\toprule
& \textbf{MathVista} & \textbf{KoNET Math} & \textbf{KCSAT-ML (Ours)} \\
\midrule
\#Item (core / total) & 1,000 / 6,141 & 24 & 339 / 664 \\
Year & - & 2024 & 2014--2025 \\
\rowcolor{bluee!20} Human Statistics & \xmark & \cmark (21 items) & \textbf{\cmark (339 items)} \\
\bottomrule
\end{tabular}
\end{adjustbox}
\caption{\label{tab:benchmark_comparison} \textbf{Comparison with existing math benchmarks.} KCSAT-ML provides nationwide-cohort error rates on 339 items, expanding KoNET's KCSAT subset (21 with human statistics) by $16\times$.}
\end{table}

\paragraph{Difficulty tiers.}
Table~\ref{tab:benchmark_comparison} compares KCSAT-ML against existing math benchmarks. All 339 core items are 3- or 4-point questions\footnote{Items are scored 2, 3, or 4 points; error rates are released only for the 3- and 4-point bands.} (4-point: 263; 3-point: 76), each typically requiring multi-step reasoning. Using official human error rates rather than point value alone, we bin the 339 core items into three tiers: \textit{Easy} (0--50\%: 148 items), \textit{Medium} (51--75\%: 113 items), and \textit{Hard} (76--100\%: 78 items).

\subsection{Models and TTS Conditions}
We evaluate open-weight LLMs/VLMs and closed-source APIs on the KCSAT-ML core set. VLMs receive the problem image directly, while LLMs are provided OCR text. We report \textit{Score} (accuracy, \%) and \textit{Avg.Tok} (average generated output tokens per item) in Table~\ref{tab:main_results}.

\paragraph{Answer evaluation.}
Core items are reformulated into a short-answer format to remove multiple-choice guessing while preserving original content and layout. Predictions are scored with an LLM-based equivalence judge (\texttt{GPT-5-2025-08-07}) that performs a narrow equivalence check between the predicted final answer and the reference (e.g., whether $\frac{1}{2}$ and $0.5$ denote the same value) and outputs \texttt{Correct}/\texttt{Incorrect}. The judge inspects the model's full response but evaluates only the equivalence of the extracted final answer against the fixed ground-truth, not the logical correctness of the reasoning chain; it is also blind to model identity and TTS configuration, so judge-induced bias across systems should be small. Full prompt templates appear in Figure~\ref{fig:prompt}.

\paragraph{Inference settings (wo.\,TTS vs.\ w.\,TTS).}
We compare (i) a low-budget, answer-only baseline (\textit{wo.\,TTS}) against (ii) higher-budget settings (\textit{w.\,TTS}) that allocate additional inference-time compute, either by adjusting inference settings of the same checkpoint or by using a reasoning-optimised variant where available. Only the latter changes the weights; Appendix~\ref{sec:tts_taxonomy} enumerates which families fall under each of the four intervention types and re-derives the main claims on checkpoint-fixed comparisons alone.

\begin{figure}[t!]
  \centering
  \includegraphics[width=\linewidth]{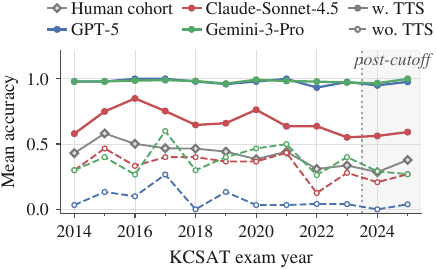}
  \caption{\textbf{Cross-model temporal accuracy on KCSAT-ML.} Per-year mean accuracy under wo.\,TTS (dashed) and w.\,TTS (solid) for GPT-5, Claude-Sonnet-4.5, and Gemini-3-Pro (2014--2025); grey diamonds: human cohort. Shaded: post-cutoff window anchored on GPT-5.}
  \label{fig:contamination_temporal}
\end{figure}

\paragraph{Data contamination check.}
KCSAT items are publicly released, raising a reasonable contamination concern. We anchor the check on GPT-5 (cut-off $\sim$September 2024, preceding the November 2024 administration of the 2025 exam): under memorisation we would expect a drop on post-cut-off items and a shrinking wo./w.\,TTS gap on pre-cut-off ones, but observe neither, and the same trajectory appears for Claude-Sonnet-4.5 and Gemini-3-Pro (Figure~\ref{fig:contamination_temporal}). Full evidence is detailed in the Limitations section.

\begin{table}[t!]
\centering
\begin{adjustbox}{width=\linewidth}
\setlength{\tabcolsep}{2pt}
\begin{tabular}{lccc}
\toprule
\textbf{Model} & \textbf{TTS} & \textbf{Avg.Tok} & \textbf{Score} \\
\midrule
\multicolumn{4}{l}{\textit{\textbf{Open-Weight LLMs (with OCR)}}} \\
\midrule
\rowcolor{gray!10} Exaone-4.0-32B & & 6 & 3.8 \\
\rowcolor{bluee!20} \textbf{Exaone-4.0-32B} & \checkmark & \textbf{2,396} & \textbf{39.8} \\
\addlinespace[0.1cm]
\rowcolor{gray!10} Qwen3-30B-A3B-Instruct & & 5 & 7.4 \\
\rowcolor{bluee!10} Qwen3-30B-A3B-Instruct & \checkmark & 4,063 & 62.2 \\
\rowcolor{bluee!20} \textbf{Qwen3-30B-A3B-Thinking} & \checkmark & \textbf{7,678} & \textbf{74.0} \\
\midrule
\multicolumn{4}{l}{\textit{\textbf{Open-Weight VLMs}}} \\
\midrule
\rowcolor{gray!10} HyperCLOVA-X-Seed-Think-32B & & 746 & 39.5 \\
\rowcolor{bluee!20} \textbf{HyperCLOVA-X-Seed-Think-32B} & \checkmark & \textbf{7,431} & \textbf{87.2} \\
\addlinespace[0.1cm]
\rowcolor{gray!10} Qwen3-VL-8B-Instruct & & 58 & 6.5 \\
\rowcolor{bluee!10} Qwen3-VL-8B-Instruct & \checkmark & 5,450 & 46.3 \\
\rowcolor{bluee!20} \textbf{Qwen3-VL-8B-Thinking} & \checkmark & \textbf{8,842} & \textbf{52.2} \\
\addlinespace[0.1cm]
\rowcolor{gray!10} Qwen3-VL-32B-Instruct & & 5 & 6.5 \\
\rowcolor{bluee!10} Qwen3-VL-32B-Instruct & \checkmark & 3,703 & 54.6 \\
\rowcolor{bluee!20} \textbf{Qwen3-VL-32B-Thinking} & \checkmark & \textbf{7,808} & \textbf{61.7} \\
\midrule
\multicolumn{4}{l}{\textit{\textbf{Closed-Source APIs (Selected Baselines)}}} \\
\midrule
\addlinespace[0.1cm]
\rowcolor{gray!10} GPT-5.1 (None) & & 13 & 8.8 \\
\rowcolor{bluee!20} \textbf{GPT-5.1 (High)} & \checkmark & \textbf{6,587} & \textbf{93.2} \\
\addlinespace[0.1cm]
\rowcolor{gray!10} Claude-Opus-4.5 (Non-think) & & 7 & 13.3 \\
\rowcolor{bluee!20} \textbf{Claude-Opus-4.5 (Think)} & \checkmark & \textbf{3,617} & \textbf{87.6} \\
\addlinespace[0.1cm]
\rowcolor{gray!10} Gemini-3-Pro-Prev (Low) & & 23 & 37.4 \\
\rowcolor{bluee!20} \textbf{Gemini-3-Pro-Prev (High)} & \checkmark & \textbf{6,913} & \textbf{99.4} \\
\bottomrule
\end{tabular}
\end{adjustbox}
\caption{\textbf{Selected KCSAT-ML Results.} Full results in Appendix Table~\ref{tab:full_results}. Baselines are shaded gray, their TTS counterparts blue.}
\label{tab:main_results} 
\end{table}

\begin{figure*}[t!]
  \centering
  \includegraphics[width=\linewidth]{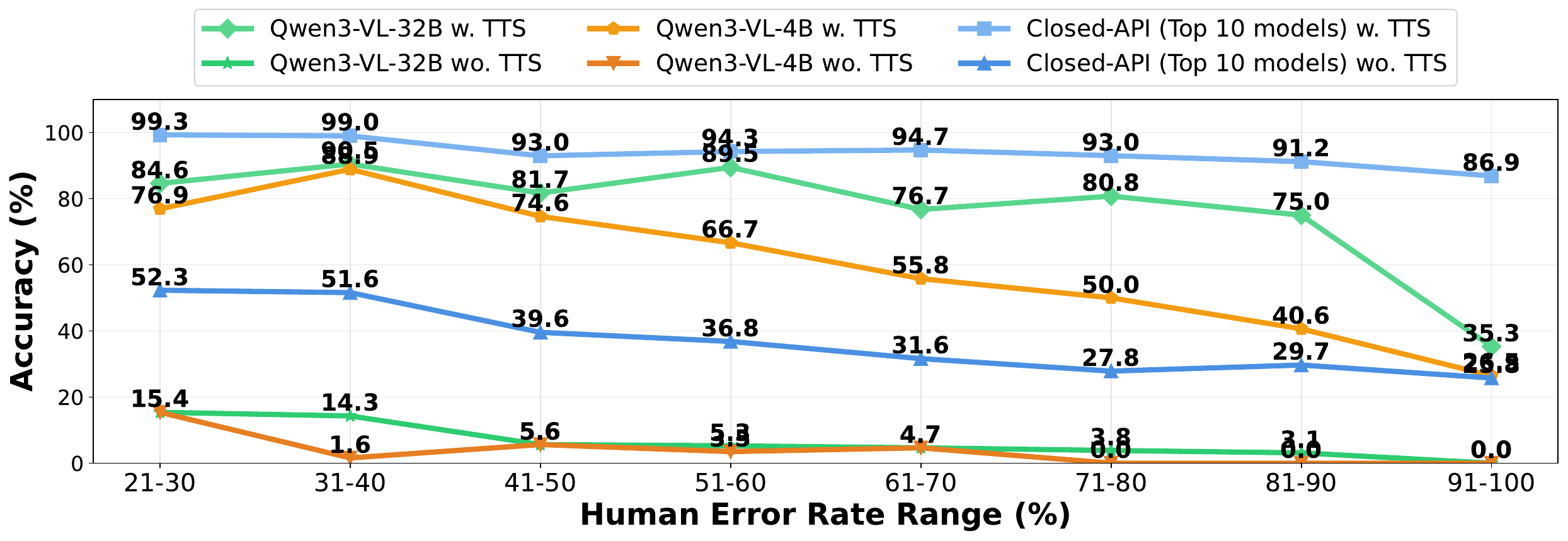}
\caption{\textbf{Model scaling vs.\ test-time scaling (TTS) by human difficulty.} KCSAT-ML accuracy across human error-rate bins for Qwen3-VL (4B vs.\ 32B) and the top-10 closed APIs, with and without TTS.}
  \label{fig:model_vs_tts}
\end{figure*}

\begin{figure}[t!]
  \centering
  \includegraphics[width=\linewidth]{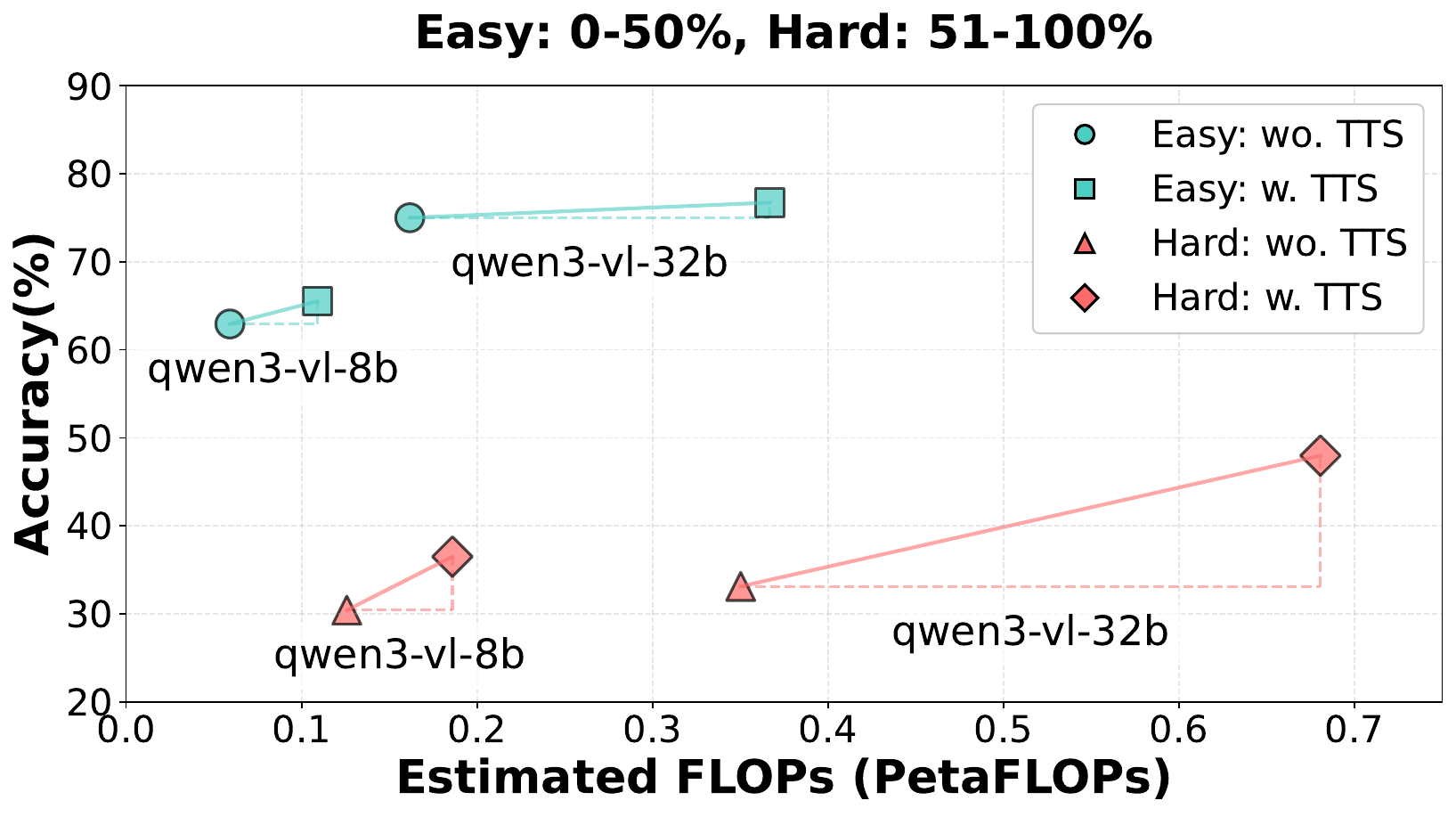}
    \caption{\textbf{When scaling helps: Qwen3-VL-8B vs.\ 32B.} Accuracy vs.\ estimated FLOPs on Easy and Hard subsets (split at $50\%$ cohort error).}
  \label{fig:tts_helps}
\end{figure}

\begin{figure}[t!]
  \centering
  \includegraphics[width=\linewidth]{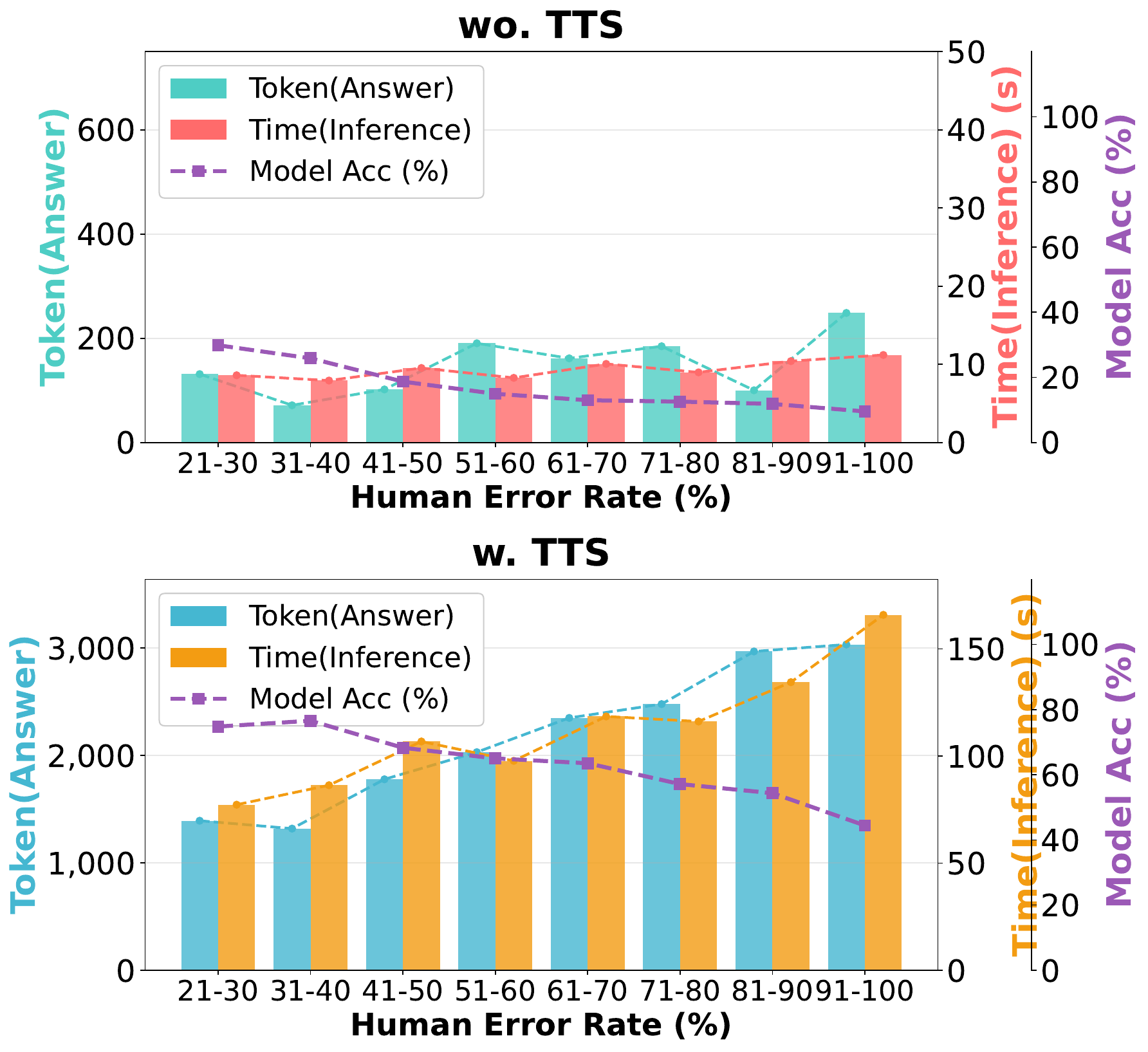}
    \caption{\textbf{Cost--accuracy trade-off of TTS.} Output tokens (blue), inference time (orange), and accuracy (purple) by human error rate, under wo.\,TTS (top) and w.\,TTS (bottom).}
  \label{fig:tts_cost}
\end{figure}

\section{Analysis and Findings}
\label{sec:results}

\begin{table}[t!]
\centering
\setlength{\tabcolsep}{4pt}
\renewcommand{\arraystretch}{1.12}
\begin{adjustbox}{max width=\linewidth}
\begin{tabular}{lccc|ccc}
\toprule
& \multicolumn{3}{c}{\textbf{MathVista}}
& \multicolumn{3}{c}{\textbf{KCSAT-ML}} \\
\cmidrule(lr){2-4}\cmidrule(lr){5-7}
\textbf{Model}
& \textbf{wo.\ TTS} & \textbf{w.\ TTS} & \textbf{$\Delta$}
& \textbf{wo.\ TTS} & \textbf{w.\ TTS} & \textbf{$\Delta$} \\
\midrule
Gemini-2.5-Flash & 75.3 & 79.4 & +4.1  & 11.1 & 86.7 & +75.6 \\
GPT-5-Mini       & 59.6 & 79.1 & +19.5 & 5.0  & 86.1 & +81.1 \\
Qwen3-VL-32B      & 83.8 & 85.9 & +2.1  & 6.5  & 61.7 & +55.2 \\
\bottomrule
\end{tabular}
\end{adjustbox}
\caption{\textbf{Effect of Test-Time Scaling (TTS).} wo.\,TTS: each model's default decoding; w.\,TTS: the official reasoning configuration of the same family. MathVista numbers are on \textit{testmini}. The wo.\,TTS floor differs across benchmarks (high on MathVista, near zero on KCSAT-ML), which partly accounts for the $\Delta$ disparity.}
\label{tab:vs_mathvista}
\end{table}

\subsection{Overall Impact of Test-Time Scaling}
Table~\ref{tab:main_results} reveals a clear \emph{collapse--recovery} pattern: most models collapse to single-digit accuracy under low decoding budgets and recover to $60$--$99\%$ once TTS is enabled (e.g., HyperCLOVA-X-Seed-Think-32B $39.5\to87.2$, Gemini-3-Pro-Preview reaches $99.4$). The wo./w.\,TTS gap on KCSAT-ML far exceeds that on MathVista (Table~\ref{tab:vs_mathvista}), leaving substantial room for inference-time compute to matter and making KCSAT-ML a sharper instrument for studying when reasoning budgets are necessary.

\subsection{Difficulty-Conditioned Scaling: Parameters vs.\ Test-Time Budget}
Figure~\ref{fig:model_vs_tts} shows where the recovery happens. Stratified by official human error-rate bins, accuracy falls off abruptly as human error rises under the low-budget setting (parameter scaling mainly helps the easier bins), while TTS reverses this pattern on the Hard tier ($\geq 76\%$ human error), so an open-weight VLM with TTS (e.g., Qwen3-VL-32B) can outperform the average closed-source API under low budgets. Figure~\ref{fig:tts_helps} drills further into Qwen3-VL 8B vs.\ 32B: on easier items, parameter scaling dominates and even aggressive TTS on the 8B model does not catch up; on harder items the picture inverts, with TTS producing a steeper gain than scaling 8B to 32B under \textit{wo.\,TTS}. Beyond moderate difficulty, inference budget therefore dominates parameter scaling (Hard-tier ranking is robust across $k$-fold splits; Appendix~\ref{sec:robustness}).
\finding{1}{Parameter scaling helps Easy items, while TTS dominates Hard ones: a $4{\times}$ smaller w.\,TTS variant can beat a larger wo.\,TTS sibling on the hardest items.}

\subsection{Cost--Accuracy Trade-off}
\label{sec:cost_accuracy}
Figure~\ref{fig:tts_cost} adds the cost axis. Under the low-budget setting, output length and latency are roughly constant across difficulty bins even as accuracy falls. Under TTS, output tokens grow roughly linearly with cohort error rate (Pearson $r{=}0.59$, $p{<}10^{-30}$; slope $\approx 102$ tokens per pp of cohort error): models spend more compute on items humans find hardest. The accuracy gain, however, follows a non-monotonic curve, peaking at moderate difficulty (Sec.~\ref{sec:diff_signal}) and separating the cost of TTS from the accuracy benefit it delivers. Across closed-source models, monetary cost and accuracy are positively but imperfectly correlated ($r{=}0.45$, Spearman $\rho{=}0.66$); several mid-tier configurations with larger reasoning budgets reach both higher accuracy and lower cost than higher-tier models under low budgets, so budget allocation often outweighs model tier in practical cost-effectiveness (per-model dollar costs in Appendix~\ref{sec:appendix_detail}).
\finding{2}{TTS \emph{cost} scales linearly with human difficulty ($r{=}0.59$), but TTS \emph{gain} is non-monotonic, peaking at moderate (not maximal) difficulty.}

\subsection{Cohort Error Rate as a Stronger Difficulty Signal}
\label{sec:diff_signal}
Do KCSAT's \emph{examiner-assigned point values}, a heuristic tier annotation similar to the 1--5 levels in MATH~\cite{hendrycks2021math}, already carry the information that cohort error rates do? We compare the two by regressing per-item mean model accuracy on each signal in turn.

\begin{figure}[t!]
  \centering
  \includegraphics[width=\linewidth]{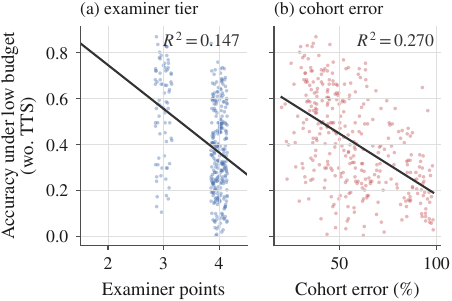}
    \caption{\textbf{Cohort error rate predicts model behaviour more strongly than examiner-assigned tiers.} Per-item mean accuracy (wo.\,TTS) vs.\ (a) examiner points and (b) cohort error rate; $N{=}339$.}
\label{fig:diff_signal}
\end{figure}

The cohort error rate is the stronger signal. It predicts per-item mean accuracy with nearly twice the explanatory power of examiner-assigned points ($R^2 = 0.27$ vs.\ $0.15$ under wo.\,TTS; $0.18$ vs.\ $0.05$ under TTS), and a joint OLS regression confirms the cohort signal carries information the heuristic tier does not (full statistics in Appendix~\ref{sec:appendix_detail}; the comparison extended to every model-independent signal in Appendix~\ref{sec:proxies}). This continuous, behaviourally grounded variation is what surfaces the non-monotonic TTS-gain curve hidden under coarse point-value tiers.
\finding{3}{Human-grounded difficulty predicts model behaviour with roughly $2{\times}$ the explanatory power of examiner-assigned tiers.}

\subsection{Two Faces of TTS: Anti-Scaling and Overthinking}
\label{sec:two_faces}
The non-monotonic gain curve from Sec.~\ref{sec:diff_signal} is not one effect: at the item level, it splits into two opposite failure modes at the difficulty extremes. We isolate them here using the per-item logs and provide a fuller catalog in Appendix~\ref{sec:edge_cases}.

\paragraph{Anti-scaling on the hardest items.}
On the 23 items with cohort error $\geq 93\%$, Qwen3-VL-32B-Instruct in force-non-think mode is majority-incorrect on every one ($0/23$), terminating with $\leq 4$ output tokens. Adding TTS to the same 32B backbone (Qwen3-VL-32B-Thinking) recovers $6/23$ items; the $4\times$ smaller Qwen3-VL-8B-Thinking under TTS recovers $9/23$, with the two TTS-on configurations overlapping on $4$ items and the rest split roughly evenly. The smaller w.\,TTS variant therefore solves items the larger wo.\,TTS sibling cannot, and the same outcome holds against Qwen3-30B-A3B-Instruct (force-non-think): the source of improvement is inference-time compute, not parameter count. Five representative items with per-trial outcomes appear in Appendix Table~\ref{tab:edge_cases_combined}(a).

\paragraph{Overthinking on easier items.}
At cohort error rates below $60\%$ (204 items: the Easy tier plus the lower portion of Medium), the converse pattern emerges \emph{within a single backbone}: TTS degrades accuracy on items the wo.\,TTS variant already solves. Counting items on which the wo.\,TTS variant is majority-correct but the w.\,TTS variant is majority-incorrect: $11/204$ ($5.4\%$) for Qwen3-VL-32B, $7/204$ ($3.4\%$) for Qwen3-30B-A3B, and $4/204$ ($2.0\%$) for GPT-OSS-120B. Each rate is modest in isolation, but the same direction appears across three families with distinct architectures and training pipelines (per-configuration decoding noise is much smaller, Appendix~\ref{sec:robustness}), so overthinking is a systematic TTS failure mode rather than a per-item or per-model artifact, echoing concurrent observations on o1-like models~\citep{chen2024donot}. Per-family lists with token counts and qualitative failure modes (e.g., abandoning textbook approaches for higher-machinery alternatives) appear in Appendix Table~\ref{tab:edge_cases_combined}(b).

Together, these patterns show TTS is difficulty-asymmetric: extra compute can be decisive on Hard items but harmful on Easy ones.
\finding{4}{Within a single family, TTS reverses direction at the difficulty extremes: \emph{anti-scaling} on Hard, \emph{overthinking} on Easy. No uniform budget serves both.}

\subsection{TTS-Induced Difficulty Alignment Across Capability Levels}
\label{sec:three_regimes}

\begin{figure}[t!]
  \centering
  \includegraphics[width=\linewidth]{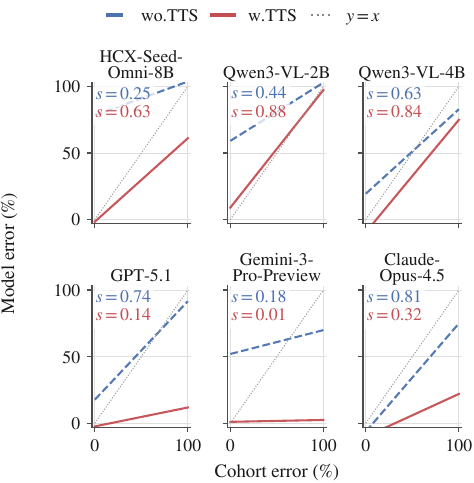}
\caption{\textbf{Three regimes of TTS difficulty alignment.} Per-item model error rate ($y$) vs.\ cohort error rate ($x$), with wo.\,TTS (dashed) and w.\,TTS (solid) regression lines; $s$ marks the fitted slope (intercepts in Appendix~\ref{sec:appendix_detail}). Top: weak/mid-tier VLMs; bottom: frontier closed-source APIs.}
  \label{fig:three_regimes}
\end{figure}

The TTS asymmetry from Section~\ref{sec:two_faces} appears at the item level within a single model. Viewed across models, TTS produces an analogous re-positioning along a single axis: the relationship between model error and human cohort error. The regression view compresses each model into two interpretable quantities. For each model with both wo.\,TTS and w.\,TTS runs we regress per-item model error rate on cohort error rate separately for each TTS condition, treating slope (failure--difficulty alignment) and intercept (offset relative to the $y{=}x$ diagonal) as the two summary statistics. The $y{=}x$ diagonal marks parity with the human cohort: the slope captures how closely model failures track human difficulty, the intercept measures vertical offset from human performance, and a regression line sitting below $y{=}x$ corresponds to error rates below the cohort across the difficulty range.

\paragraph{Three regimes.}
Figure~\ref{fig:three_regimes} reveals three regimes (per-model slopes and intercepts are tabulated in Appendix~\ref{sec:appendix_detail}). \textit{No-reasoning} (top row, wo.\,TTS): a near-flat regression sitting at high error, with failures uncorrelated with difficulty. \textit{Human-aligned} (top row w.\,TTS, bottom row wo.\,TTS): slope $\approx 1$ with positive intercept, failures proportional to human difficulty. \textit{Below-cohort} (bottom row w.\,TTS): slope $<\!0.4$ combined with a strongly negative intercept, so the model's error rate sits below the human cohort across the difficulty range. The same TTS turn-on drives opposite transitions depending on the starting point: it lifts weak models from \textit{no-reasoning} into \textit{human-aligned} (e.g., Qwen3-VL-2B's slope goes from $0.44$ to $0.88$), and frontier models from \textit{human-aligned} into \textit{below-cohort} (GPT-5.1's intercept drops to $-9.4$). This view places the anti-scaling and overthinking findings of Section~\ref{sec:two_faces} on common ground as item-level manifestations of one capability-conditional mechanism~\citep{wei2022emergent}: TTS reshapes \emph{which items} a model fails on, with direction determined by the model's starting position.
\finding{5}{The same TTS turn-on drives \emph{opposite} transitions for weak and frontier models, partitioning capability into three regimes: \emph{no-reasoning}, \emph{human-aligned}, \emph{below-cohort}.}

\subsection{Beyond Accuracy: Measuring TTS Reasoning Quality}
\label{sec:drg}

\begin{figure*}[t!]
  \centering
  \includegraphics[width=\linewidth]{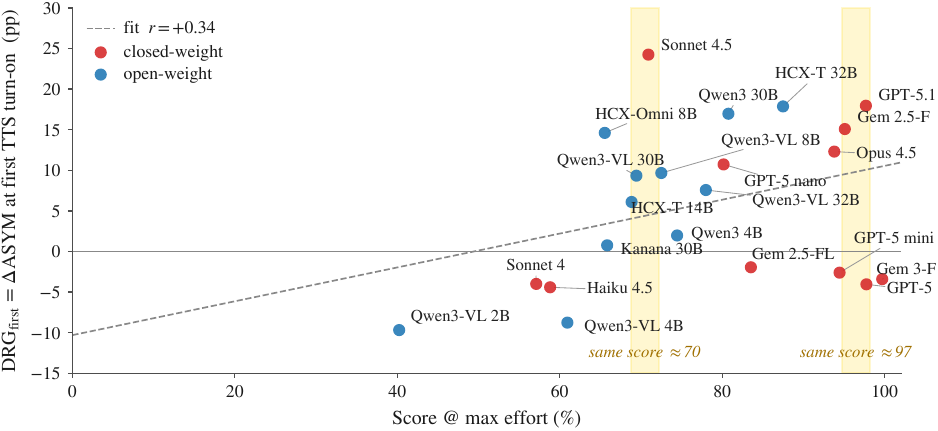}
\caption{\textbf{DRG vs.\ accuracy.} For 22 model families with both wo.\,TTS and w.\,TTS runs, accuracy at max TTS effort ($x$) against DRG at first TTS turn-on ($y$); dashed: linear fit. Vertical gold bands highlight clusters with similar accuracy but wide DRG spread. Closed-weight (red) and open-weight (blue) distribute similarly along $y$ (Mann--Whitney $p{=}0.95$).}
  \label{fig:drg_scatter}
\end{figure*}

The patterns in Sections~\ref{sec:two_faces}--\ref{sec:three_regimes} indicate that TTS reshapes \emph{which items} a model fails on, asymmetrically across difficulty. The question we now ask is one accuracy cannot answer: when TTS improves a model, does it reduce failures on the items humans found hardest, or does it mainly clean up easier residual errors?

\paragraph{Sub-cohort fraction (SCF).} For a given (model, TTS setting) configuration and a difficulty bin $b \in \{\text{Easy}, \text{Hard}\}$, let $a_b$ be the number of items in bin $b$ where the model's error rate exceeds the human cohort's, and $\bar{a}_b$ the number where it falls below. We define the \emph{sub-cohort fraction}
$$
\mathrm{SCF}_b \;=\; \frac{a_b}{a_b + \bar{a}_b}.
$$
$\mathrm{SCF}_b$ measures how often the model under-performs the human cohort within bin $b$.  We use Easy ($\le 50\%$ cohort error, $148$ items) and Hard ($\ge 76\%$, $78$ items) tiers as defined in Section~\ref{sec:benchmark_construction}.

\paragraph{Difficulty-aligned Reasoning Gain (DRG).} Turning TTS on shifts the Easy/Hard SCF asymmetry, locating where the model's residual under-performance against humans concentrates. We define
$$
\mathrm{DRG} \;=\; \bigl(\mathrm{SCF}_H^{\text{off}} - \mathrm{SCF}_H^{\text{on}}\bigr) - \bigl(\mathrm{SCF}_E^{\text{off}} - \mathrm{SCF}_E^{\text{on}}\bigr),
$$
taken at the first TTS turn-on. Positive DRG means TTS preferentially closes the Hard-bin gap relative to humans; negative DRG means the gain is concentrated on Easy items while Hard failures remain; near-zero DRG means the improvement is roughly uniform across difficulty.
\paragraph{(1) DRG is nearly orthogonal to accuracy.}  Across the $22$ families, the Pearson correlation between final score and DRG is only $r=+0.34$ (Figure~\ref{fig:drg_scatter}).  Within the high-accuracy cluster (max-effort score in $[86, 99]$), five frontier models span a $22$\,pp DRG range, and same-family generations can split sharply (e.g., Claude Sonnet-4 at $-4.0$ vs.\ Sonnet-4.5 at $+24.3$). The DRG sign decomposes the gain by difficulty tier: Sonnet-4.5's TTS budget preferentially clears above-cohort failures on the items humans found hardest, while Sonnet-4 concentrates its reduction on the Easy tier instead, so the two checkpoints reach comparable accuracy via near-opposite reasoning targets.  Per-family values are listed in Appendix~\ref{sec:appendix_detail}.
\paragraph{(2) DRG is locked in at the first TTS turn-on.}  For families with multiple TTS budgets, DRG at the lowest TTS-on setting correlates with DRG at the highest setting at $r=+0.94$.  Raising the budget lifts accuracy marginally but barely shifts DRG: where TTS places its gains along the difficulty axis is locked in at turn-on.
\paragraph{(3) Classification by DRG separates failure profiles.}  Thresholding DRG at $\pm$ a small margin yields HIGH ($\ge\!+10$\,pp; Hard-first saturation), MID ($-5$ to $+10$\,pp), and LOW ($\le\!-5$\,pp; residual weaknesses concentrated in the Hard bin) groups (8/12/2 families). The LOW group shares a recurring failure: systematic over-engineering on textbook-level items, drawing thousands of reasoning tokens on problems wo.\,TTS solves in a handful (qualitative examples in Appendix~\ref{sec:appendix_detail}).
\paragraph{Measurement versus interpretation.}
What DRG measures is a distributional fact: where a model's residual failures sit relative to the human difficulty axis, before and after TTS. Reading a positive DRG as better \emph{reasoning} is an interpretation layered on that measurement, and one this benchmark cannot adjudicate---we observe outcomes, not reasoning processes, and a model may reach a hard item's answer by a route no examinee would take. We therefore treat DRG as a diagnostic of where failures fall, and keep reasoning-quality language for statements we mark as interpretive.

\finding{6}{DRG decouples \emph{difficulty alignment} from accuracy ($r{=}0.34$) and settles at the first TTS turn-on: same-score models can distribute their failures in opposite ways.}

\section{Conclusion}
\label{sec:conclusion}
We introduce \textbf{KCSAT-ML}, a decade of KCSAT mathematics with nationwide-cohort human error rates, paired with \emph{Difficulty-aligned Reasoning Gain} (DRG): a score-orthogonal metric for measuring how well a model's failures align with the human difficulty axis.
Together, they show that test-time compute changes not only \emph{how often} but \emph{which items} models solve correctly: a difficulty-dependent collapse--recovery on the high-difficulty tail, \emph{anti-scaling} and \emph{overthinking} at opposite ends, and three capability-conditional alignment regimes across model families.
The framework extends to any examination programme releasing item-level human statistics, offering a general probe of where model failures fall against human-grounded difficulty.

\section*{Limitations}
\label{sec:limitations}
 
\paragraph{Human signal as proxy.}
Official human error rates are measured under the original KCSAT formats, while our evaluation reformulates items into a short-answer interface. The released rates therefore serve as a difficulty proxy rather than a calibrated measurement under our protocol, especially for originally multiple-choice items. Our analyses, however, depend on relative rankings across difficulty bins rather than absolute error rates, and the bin-level ordering is preserved under reformatting.
 
\paragraph{LLM judge.}
An LLM-based equivalence judge (\texttt{GPT-5-2025-08-07}; prompt in Appendix Figure~\ref{fig:prompt}) handles symbolic answers and formatting variation, but may introduce occasional judging errors (validated against a deterministic oracle in Appendix~\ref{sec:judge_validation}). The judge is blind to model identity and TTS configuration, which limits systematic per-system bias, and we report trends that are stable across difficulty bins and inference budgets to further reduce sensitivity to individual judgments.
 
\paragraph{Contamination.}
KCSAT items are publicly released, but three observations argue against simple surface-form memorization as a driver of our results: (i) per-year wo.\,TTS accuracy shows no upward shift on pre-cutoff items relative to post-cutoff (Figure~\ref{fig:contamination_temporal}); (ii) wo.\,TTS collapse to single-digit accuracy on public items would not require reasoning budget to surface if final answers were merely recalled; and (iii) the year-invariant wo./w.\,TTS gap (per-family mean $34$--$91$\,pp) is incompatible with memorized completions surfacing without inference compute. More subtle exposure --- such as memorized solution traces that surface under reasoning modes --- remains possible, and evaluation on newly created or access-controlled items is a complementary direction rather than a prerequisite for the present conclusions.
 
\paragraph{TTS heterogeneity across model families.}
The wo./w.\,TTS contrast is not an identical intervention across model families: for some it changes only the inference-time budget or reasoning mode of a fixed checkpoint, for others it uses a reasoning-optimized checkpoint or vendor-provided reasoning mode. We treat this as a feature rather than a bug --- it reflects how TTS is exposed in current model APIs and releases, and is the practical setting in which users choose inference budgets. Strictly causal same-checkpoint conclusions are restricted to model families where such ablations are available (per-family taxonomy and same-checkpoint re-analysis in Appendix~\ref{sec:tts_taxonomy}); the broader difficulty-conditioned patterns we report do not require them.

\bibliography{references,anthology-1,anthology-2}

\begin{thebibliography}{29}
\providecommand{\natexlab}[1]{#1}

\bibitem[{Benedetto et~al.(2023)Benedetto, Cremonesi, Caines, Buttery,
  Cappelli, Giussani, and Turrin}]{benedetto2023survey}
Luca Benedetto, Paolo Cremonesi, Andrew Caines, Paula Buttery, Andrea Cappelli,
  Andrea Giussani, and Roberto Turrin. 2023.
\newblock \href {https://doi.org/10.1145/3556538} {{A Survey on Recent
  Approaches to Question Difficulty Estimation from Text}}.
\newblock \emph{ACM Computing Surveys}, 55(9):1--37.

\bibitem[{Bi et~al.(2025)Bi, Han, Liu, Tang, and Wang}]{bi2024forestofthought}
Zhenni Bi, Kai Han, Chuanjian Liu, Yehui Tang, and Yunhe Wang. 2025.
\newblock \href {https://doi.org/10.48550/arXiv.2412.09078}
  {{Forest-of-Thought: Scaling Test-Time Compute for Enhancing LLM Reasoning}}.
\newblock In \emph{Proceedings of the 42nd International Conference on Machine
  Learning (ICML)}, volume 267 of \emph{Proceedings of Machine Learning
  Research}, pages 4253--4267.

\bibitem[{Chen et~al.(2025)Chen, Yu, Wang, Hu, Zhang, Wang, Tang, Xiong, Li,
  Wang, Yang, and Li}]{chen2025xverify}
Ding Chen, Qingchen Yu, Pengyuan Wang, Mengting Hu, Wentao Zhang, Zhengren
  Wang, Bo~Tang, Feiyu Xiong, Xinchi Li, Chao Wang, Minchuan Yang, and Zhiyu
  Li. 2025.
\newblock {xVerify: Efficient Answer Verifier for Reasoning Model Evaluations}.
\newblock \emph{arXiv preprint arXiv:2504.10481}.

\bibitem[{Chen et~al.(2024)Chen, Xu, Liang, He, Pang, Yu, Song, Liu, Zhou,
  Zhang, Wang, Tu, Mi, and Yu}]{chen2024donot}
Xingyu Chen, Jiahao Xu, Tian Liang, Zhiwei He, Jianhui Pang, Dian Yu, Linfeng
  Song, Qiuzhi Liu, Mengfei Zhou, Zhuosheng Zhang, Rui Wang, Zhaopeng Tu,
  Haitao Mi, and Dong Yu. 2024.
\newblock {Do Not Think That Much for 2+3=? On the Overthinking of o1-Like
  LLMs}.
\newblock \emph{arXiv preprint arXiv:2412.21187}.

\bibitem[{Embretson and Reise(2000)}]{embretson2000irt}
Susan~E. Embretson and Steven~P. Reise. 2000.
\newblock \emph{{Item Response Theory for Psychologists}}.
\newblock Lawrence Erlbaum Associates, Mahwah, NJ.

\bibitem[{Gao et~al.(2025)Gao, Song, Yang, Cai, Miao, Dong, Li, Ma, Chen, Xu,
  Tang, Wang, Zan, Quan, Zhang, Sha, Zhang, Ren, Liu, and
  Chang}]{gao2024omnimath}
Bofei Gao, Feifan Song, Zhe Yang, Zefan Cai, Yibo Miao, Qingxiu Dong, Lei Li,
  Chenghao Ma, Liang Chen, Runxin Xu, Zhengyang Tang, Benyou Wang, Daoguang
  Zan, Shanghaoran Quan, Ge~Zhang, Lei Sha, Yichang Zhang, Xuancheng Ren,
  Tianyu Liu, and Baobao Chang. 2025.
\newblock \href {https://doi.org/10.48550/arXiv.2410.07985} {{Omni-MATH: A
  Universal Olympiad Level Mathematic Benchmark For Large Language Models}}.
\newblock In \emph{International Conference on Learning Representations}.

\bibitem[{Guo et~al.(2025)Guo, Yang, Zhang, Song, Wang, Zhu, Xu, Zhang, Ma, Bi,
  and {DeepSeek-AI Team}}]{deepseek2025r1}
Daya Guo, Dejian Yang, Haowei Zhang, Junxiao Song, Peiyi Wang, Qihao Zhu,
  Runxin Xu, Ruoyu Zhang, Shirong Ma, Xiao Bi, and {DeepSeek-AI Team}. 2025.
\newblock \href {https://doi.org/10.1038/s41586-025-09422-z} {{DeepSeek-R1
  Incentivizes Reasoning in LLMs through Reinforcement Learning}}.
\newblock \emph{Nature}, 645(8081):633--638.

\bibitem[{He et~al.(2024)He, Luo, Bai, Hu, Thai, Shen, Hu, Han, Huang, Zhang,
  Liu, Qi, Liu, and Sun}]{he-etal-2024-olympiadbench}
Chaoqun He, Renjie Luo, Yuzhuo Bai, Shengding Hu, Zhen Thai, Junhao Shen, Jinyi
  Hu, Xu~Han, Yujie Huang, Yuxiang Zhang, Jie Liu, Lei Qi, Zhiyuan Liu, and
  Maosong Sun. 2024.
\newblock \href {https://doi.org/10.18653/v1/2024.acl-long.211}
  {{O}lympiad{B}ench: A challenging benchmark for promoting {AGI} with
  olympiad-level bilingual multimodal scientific problems}.
\newblock In \emph{Proceedings of the 62nd Annual Meeting of the Association
  for Computational Linguistics (Volume 1: Long Papers)}, pages 3828--3850,
  Bangkok, Thailand. Association for Computational Linguistics.

\bibitem[{Hendrycks et~al.(2021)Hendrycks, Burns, Kadavath, Arora, Basart,
  Tang, Song, and Steinhardt}]{hendrycks2021math}
Dan Hendrycks, Collin Burns, Saurav Kadavath, Akul Arora, Steven Basart, Eric
  Tang, Dawn Song, and Jacob Steinhardt. 2021.
\newblock {Measuring Mathematical Problem Solving With the MATH Dataset}.
\newblock In \emph{Advances in Neural Information Processing Systems (NeurIPS)
  Datasets and Benchmarks Track}.

\bibitem[{Lalor et~al.(2016)Lalor, Wu, and Yu}]{lalor-etal-2016-building}
John~P. Lalor, Hao Wu, and Hong Yu. 2016.
\newblock \href {https://doi.org/10.18653/v1/D16-1062} {Building an evaluation
  scale using item response theory}.
\newblock In \emph{Proceedings of the 2016 Conference on Empirical Methods in
  Natural Language Processing}, pages 648--657, Austin, Texas. Association for
  Computational Linguistics.

\bibitem[{Lalor et~al.(2019)Lalor, Wu, and Yu}]{lalor-etal-2019-learning}
John~P. Lalor, Hao Wu, and Hong Yu. 2019.
\newblock \href {https://doi.org/10.18653/v1/D19-1434} {Learning latent
  parameters without human response patterns: Item response theory with
  artificial crowds}.
\newblock In \emph{Proceedings of the 2019 Conference on Empirical Methods in
  Natural Language Processing and the 9th International Joint Conference on
  Natural Language Processing (EMNLP-IJCNLP)}, pages 4249--4259, Hong Kong,
  China. Association for Computational Linguistics.

\bibitem[{Liao et~al.(2025)Liao, Xu, Dong, Li, Monz, Savarese, Sahoo, and
  Xiong}]{liao2025rewardguided}
Baohao Liao, Yuhui Xu, Hanze Dong, Junnan Li, Christof Monz, Silvio Savarese,
  Doyen Sahoo, and Caiming Xiong. 2025.
\newblock \href {https://doi.org/10.48550/arXiv.2501.19324} {{Reward-Guided
  Speculative Decoding for Efficient LLM Reasoning}}.
\newblock In \emph{International Conference on Machine Learning}.

\bibitem[{Lightman et~al.(2024)Lightman, Kosaraju, Burda, Edwards, Baker, Lee,
  Leike, Schulman, Sutskever, and Cobbe}]{lightman2024verify}
Hunter Lightman, Vineet Kosaraju, Yura Burda, Harri Edwards, Bowen Baker, Teddy
  Lee, Jan Leike, John Schulman, Ilya Sutskever, and Karl Cobbe. 2024.
\newblock {Let's Verify Step by Step}.
\newblock In \emph{International Conference on Learning Representations}.

\bibitem[{Liu et~al.(2025)Liu, Liu, Liu, Xiao, Gao, Lyu, Gu, Zhang, Wong,
  Zhang, and Chen}]{liu-etal-2025-compassverifier}
Shudong Liu, Hongwei Liu, Junnan Liu, Linchen Xiao, Songyang Gao, Chengqi Lyu,
  Yuzhe Gu, Wenwei Zhang, Derek~F. Wong, Songyang Zhang, and Kai Chen. 2025.
\newblock \href {https://doi.org/10.18653/v1/2025.emnlp-main.1698}
  {{C}ompass{V}erifier: A unified and robust verifier for {LLM}s evaluation and
  outcome reward}.
\newblock In \emph{Proceedings of the 2025 Conference on Empirical Methods in
  Natural Language Processing}, pages 33454--33482, Suzhou, China. Association
  for Computational Linguistics.

\bibitem[{Lord and Novick(1968)}]{lord1968statistical}
Frederic~M. Lord and Melvin~R. Novick. 1968.
\newblock \emph{{Statistical Theories of Mental Test Scores}}.
\newblock Addison-Wesley, Reading, MA.

\bibitem[{Lu et~al.(2024)Lu, Bansal, Xia, Liu, Li, Hajishirzi, Cheng, Chang,
  Galley, and Gao}]{lu2023mathvista}
Pan Lu, Hritik Bansal, Tony Xia, Jiacheng Liu, Chunyuan Li, Hannaneh
  Hajishirzi, Hao Cheng, Kai-Wei Chang, Michel Galley, and Jianfeng Gao. 2024.
\newblock {MathVista: Evaluating Mathematical Reasoning of Foundation Models in
  Visual Contexts}.
\newblock In \emph{International Conference on Learning Representations}.

\bibitem[{Maia~Polo et~al.(2024)Maia~Polo, Weber, Choshen, Sun, Xu, and
  Yurochkin}]{maiapolo2024tinybenchmarks}
Felipe Maia~Polo, Lucas Weber, Leshem Choshen, Yuekai Sun, Gongjun Xu, and
  Mikhail Yurochkin. 2024.
\newblock \href {https://proceedings.mlr.press/v235/maia-polo24a.html}
  {{tinyBenchmarks: Evaluating LLMs with Fewer Examples}}.
\newblock In \emph{Proceedings of the 41st International Conference on Machine
  Learning (ICML)}, volume 235 of \emph{Proceedings of Machine Learning
  Research}, pages 34303--34326.

\bibitem[{Mart{\'i}nez-Plumed et~al.(2019)Mart{\'i}nez-Plumed, Prud{\^e}ncio,
  Mart{\'i}nez-Us{\'o}, and Hern{\'a}ndez-Orallo}]{martinezplumed2019irt}
Fernando Mart{\'i}nez-Plumed, Ricardo B.~C. Prud{\^e}ncio, Adolfo
  Mart{\'i}nez-Us{\'o}, and Jos{\'e} Hern{\'a}ndez-Orallo. 2019.
\newblock \href {https://doi.org/10.1016/j.artint.2018.09.004} {{Item Response
  Theory in AI: Analysing Machine Learning Classifiers at the Instance Level}}.
\newblock \emph{Artificial Intelligence}, 271:18--42.

\bibitem[{McKenzie et~al.(2023)McKenzie, Lyzhov, Pieler, Parrish, Mueller,
  Prabhu, McLean, Kirtland, Ross, Liu, and {others}}]{mckenzie2023inverse}
Ian~R. McKenzie, Alexander Lyzhov, Michael Pieler, Alicia Parrish, Aaron
  Mueller, Ameya Prabhu, Euan McLean, Aaron Kirtland, Alexis Ross, Alisa Liu,
  and {others}. 2023.
\newblock {Inverse Scaling: When Bigger Isn't Better}.
\newblock \emph{Transactions on Machine Learning Research (TMLR)}.

\bibitem[{Muennighoff et~al.(2025)Muennighoff, Yang, Shi, Li, Fei-Fei,
  Hajishirzi, Zettlemoyer, Liang, Candes, and
  Hashimoto}]{muennighoff-etal-2025-s1}
Niklas Muennighoff, Zitong Yang, Weijia Shi, Xiang~Lisa Li, Li~Fei-Fei,
  Hannaneh Hajishirzi, Luke Zettlemoyer, Percy Liang, Emmanuel Candes, and
  Tatsunori Hashimoto. 2025.
\newblock \href {https://doi.org/10.18653/v1/2025.emnlp-main.1025} {s1: Simple
  test-time scaling}.
\newblock In \emph{Proceedings of the 2025 Conference on Empirical Methods in
  Natural Language Processing}, pages 20286--20332, Suzhou, China. Association
  for Computational Linguistics.

\bibitem[{Park and Kim(2025)}]{park-kim-2025-evaluating}
Sanghee Park and Geewook Kim. 2025.
\newblock \href {https://doi.org/10.18653/v1/2025.naacl-short.56} {Evaluating
  multimodal generative {AI} with {K}orean educational standards}.
\newblock In \emph{Proceedings of the 2025 Conference of the Nations of the
  Americas Chapter of the Association for Computational Linguistics: Human
  Language Technologies (Volume 2: Short Papers)}, pages 671--688, Albuquerque,
  New Mexico. Association for Computational Linguistics.

\bibitem[{Qiao et~al.(2025)Qiao, Tan, Dong, Wu, Sun, Song, Wang, GongQue, Lei,
  Zhang, Wei, Zhang, Qiao, Zong, Xu, Yang, Bao, Diao, Li, and
  Zhang}]{qiao2025wemath}
Runqi Qiao, Qiuna Tan, Guanting Dong, Minhui Wu, Chong Sun, Xiaoshuai Song,
  Jiapeng Wang, Zhuoma GongQue, Shanglin Lei, YiFan Zhang, Zhe Wei, Miaoxuan
  Zhang, Runfeng Qiao, Xiao Zong, Yida Xu, Peiqing Yang, Zhimin Bao, Muxi Diao,
  Chen Li, and Honggang Zhang. 2025.
\newblock {We-Math: Does Your Large Multimodal Model Achieve Human-like
  Mathematical Reasoning?}
\newblock In \emph{Proceedings of the 63rd Annual Meeting of the Association
  for Computational Linguistics (Volume 1: Long Papers)}, pages 20023--20070.

\bibitem[{Rodriguez et~al.(2021)Rodriguez, Barrow, Hoyle, Lalor, Jia, and
  Boyd-Graber}]{rodriguez-etal-2021-evaluation}
Pedro Rodriguez, Joe Barrow, Alexander Hoyle, John~P. Lalor, Robin Jia, and
  Jordan Boyd-Graber. 2021.
\newblock \href {https://doi.org/10.18653/v1/2021.acl-long.346} {Evaluation
  examples are not equally informative: How should that change {NLP}
  leaderboards?}
\newblock In \emph{Proceedings of the 59th Annual Meeting of the Association
  for Computational Linguistics and the 11th International Joint Conference on
  Natural Language Processing (Volume 1: Long Papers)}, pages 4486--4503,
  Online. Association for Computational Linguistics.

\bibitem[{Snell et~al.(2025)Snell, Lee, Xu, and Kumar}]{snell2024scaling}
Charlie Snell, Jaehoon Lee, Kelvin Xu, and Aviral Kumar. 2025.
\newblock \href {https://doi.org/10.48550/arXiv.2408.03314} {{Scaling LLM
  Test-Time Compute Optimally can be More Effective than Scaling Model
  Parameters}}.
\newblock In \emph{International Conference on Learning Representations}.

\bibitem[{Son et~al.(2026)Son, Kim, Arnett, Ko, Lee, Kang, and
  {others}}]{son2026soohak}
Guijin Son, Seungone Kim, Catherine Arnett, Hyunwoo Ko, Hyein Lee, Hyeonah
  Kang, and {others}. 2026.
\newblock {Soohak: A Mathematician-Curated Benchmark for Evaluating
  Research-level Math Capabilities of LLMs}.
\newblock \emph{arXiv preprint arXiv:2605.09063}.

\bibitem[{Wang et~al.(2023)Wang, Wei, Schuurmans, Le, Chi, Narang, Chowdhery,
  and Zhou}]{wang2022selfconsistency}
Xuezhi Wang, Jason Wei, Dale Schuurmans, Quoc~V. Le, Ed~H. Chi, Sharan Narang,
  Aakanksha Chowdhery, and Denny Zhou. 2023.
\newblock {Self-Consistency Improves Chain of Thought Reasoning in Language
  Models}.
\newblock In \emph{International Conference on Learning Representations}.

\bibitem[{Wei et~al.(2022{\natexlab{a}})Wei, Tay, Bommasani, Raffel, Zoph,
  Borgeaud, Yogatama, Bosma, Zhou, Metzler, Chi, Hashimoto, Vinyals, Liang,
  Dean, and Fedus}]{wei2022emergent}
Jason Wei, Yi~Tay, Rishi Bommasani, Colin Raffel, Barret Zoph, Sebastian
  Borgeaud, Dani Yogatama, Maarten Bosma, Denny Zhou, Donald Metzler, Ed~H.
  Chi, Tatsunori Hashimoto, Oriol Vinyals, Percy Liang, Jeff Dean, and William
  Fedus. 2022{\natexlab{a}}.
\newblock {Emergent Abilities of Large Language Models}.
\newblock \emph{Transactions on Machine Learning Research}.

\bibitem[{Wei et~al.(2022{\natexlab{b}})Wei, Wang, Schuurmans, Bosma, Ichter,
  Xia, Chi, Le, and Zhou}]{wei2022chain}
Jason Wei, Xuezhi Wang, Dale Schuurmans, Maarten Bosma, Brian Ichter, Fei Xia,
  Ed~H. Chi, Quoc~V. Le, and Denny Zhou. 2022{\natexlab{b}}.
\newblock {Chain of Thought Prompting Elicits Reasoning in Large Language
  Models}.
\newblock In \emph{Advances in Neural Information Processing Systems
  (NeurIPS)}.

\bibitem[{Zhang et~al.(2025)Zhang, Lin, Hou, Feng, and
  Li}]{zhang-etal-2025-adaptthink}
Jiajie Zhang, Nianyi Lin, Lei Hou, Ling Feng, and Juanzi Li. 2025.
\newblock \href {https://doi.org/10.18653/v1/2025.emnlp-main.184}
  {{A}dapt{T}hink: Reasoning models can learn when to think}.
\newblock In \emph{Proceedings of the 2025 Conference on Empirical Methods in
  Natural Language Processing}, pages 3716--3730, Suzhou, China. Association
  for Computational Linguistics.

\end{thebibliography}

\appendix

\section{Dataset Construction and Release}
\label{sec:dataset_construction}

The per-item error rates are published by EBSi, the educational platform operated by EBS, Korea's national public educational broadcaster. Immediately after each KCSAT administration EBSi collects the responses of a large share of the actual test takers and releases per-item statistics for the 3- and 4-point items. The exam administrator, KICE, publishes aggregate score distributions but as a matter of policy does not release item-level figures such as error rates, so the EBSi statistics are the de facto standard reference for KCSAT item difficulty in Korea, and are what we mean by \emph{official} throughout this paper. We match each published rate to its item by exam year, form, and question number; the matching procedure ships unchanged in the released dataset builder.

We publicly release the full pipeline for constructing KCSAT-ML at \url{https://github.com/naver-ai/KCSAT-ML}, along with the inference, evaluation, and analysis code used in this paper. The KCSAT problems are publicly available on the official KICE website, and we use them solely for noncommercial academic research. We do not redistribute the original exam materials; the released generator reconstructs the benchmark from these public sources using our bounding-box annotations (Figure~\ref{fig:metadata_example}).

\section{Additional Results}
\label{sec:additional_results}

This section provides additional qualitative examples, dataset statistics, full model results, and auxiliary analyses that support the main findings in Sec.~\ref{sec:results}.

\paragraph{Qualitative examples.}
Figures~\ref{fig:geometry_example} and~\ref{fig:kcsat_algebra_example} present two representative items from the KCSAT-ML core set. These examples highlight two common sources of difficulty in KCSAT mathematics: (i) dense symbolic expressions embedded within the original exam layout, and (ii) diagram-dependent reasoning that requires multi-step geometric or algebraic manipulation. For readability, English translations are provided in the figures; however, all evaluations in this paper are performed using the original Korean problem statements.

\paragraph{Dataset statistics and standardisation.}
Table~\ref{tab:kcsatml_statistics} summarises the composition of KCSAT-ML. The core subset ($N=339$) is dominated by high-point items (4 points) and spans diverse answer types, including integers, fractions, and symbolic expressions. This diversity is why we score with an answer-equivalence judge rather than brittle string matching. To reduce multiple-choice guessing and keep the evaluation interface consistent, items with available human statistics are reformulated into a short-answer format while preserving the original exam content and layout.

\paragraph{Full results across models and budgets.}
Sec.~\ref{sec:benchmark_construction} reports a selected subset of results in Table~\ref{tab:main_results}; Table~\ref{tab:full_results} reports the complete results across all evaluated open-weight models and closed-source APIs. For several model families we include multiple budget configurations (e.g., vendor-provided budget tiers), giving a finer-grained view of the cost--accuracy trade-off than the binary wo./w.\,TTS setting alone.

\begin{figure*}[t!]
  \centering
  \includegraphics[width=0.9\linewidth]{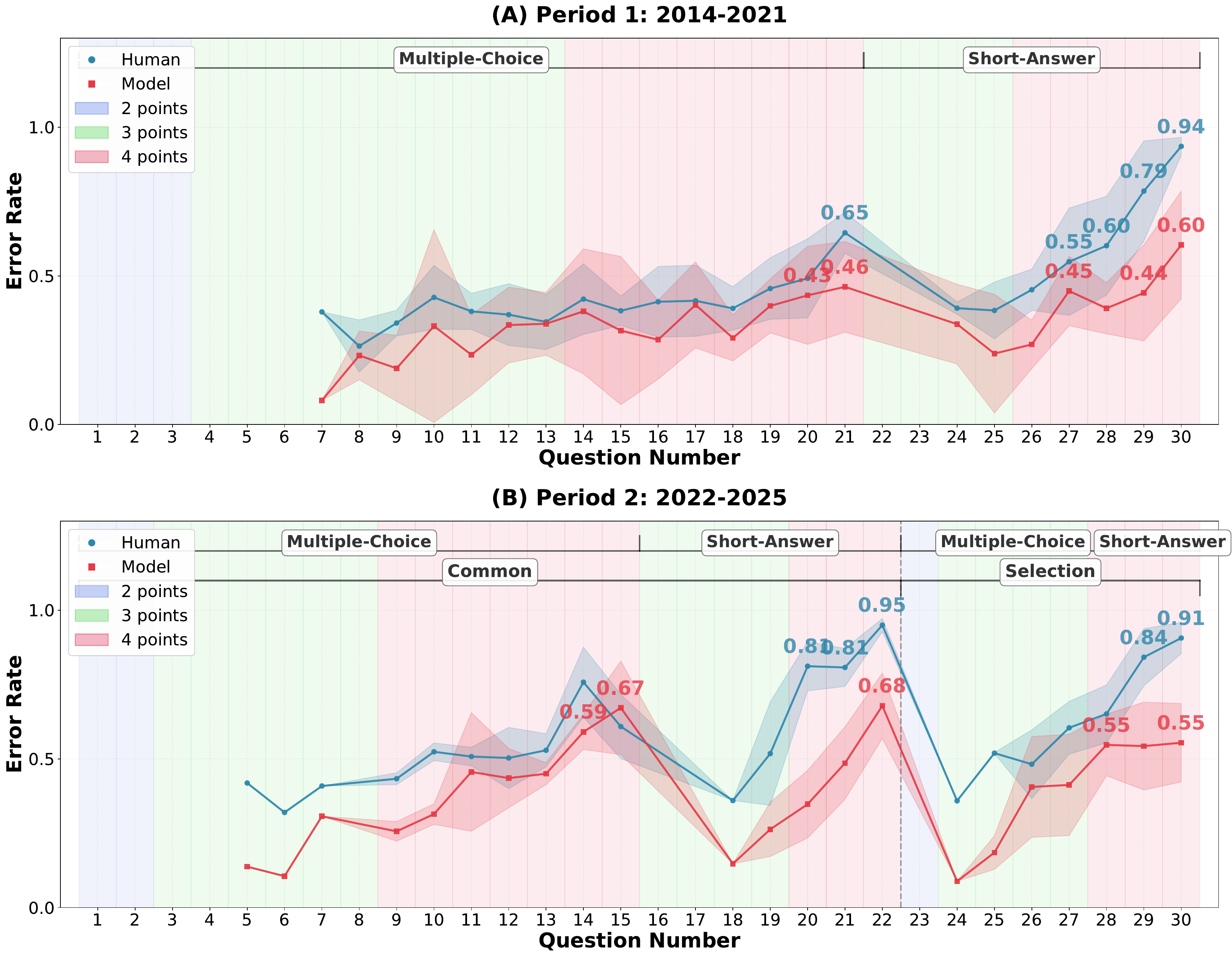}
        \caption{\textbf{Human vs.\ model error rates across question numbers.} Error rates over 30 questions in KCSAT-ML, grouped by question format (multiple-choice vs.\ short-answer) and period: (A) 2014--2021 and (B) 2022--2025. Periods are split by the 2022 introduction of the common-plus-elective structure.}
  \label{fig:numbers_chart}
\end{figure*}

\paragraph{Positional analysis (sanity check for public items).}
Because KCSAT items are publicly released, potential exposure to training data cannot be fully ruled out. Figure~\ref{fig:numbers_chart} compares human and model error rates by question number (i.e., position within an exam form) and does not show a clear monotonic trend that would suggest older items are systematically easier for the evaluated models. The temporal (year-by-year) view is given by the cross-model contamination check in Figure~\ref{fig:contamination_temporal}.

\paragraph{Prompt templates and evaluation pipeline.}
Figure~\ref{fig:prompt} shows the exact prompt templates used throughout our experiments. The wo.\,TTS setting uses a low-budget, answer-only prompt; the w.\,TTS setting uses a higher-budget prompt that permits intermediate reasoning. The figure also includes the evaluation prompt used for LLM-based answer verification (Sec.~\ref{sec:benchmark_construction}), which is needed for robust scoring of symbolic answers.

\paragraph{Structured metadata, annotation, and reconstruction.}
Figure~\ref{fig:metadata_example} shows an example of the structured metadata in KCSAT-ML, including item attributes (year, domain, question number, points), answer fields, and official human performance statistics. The figure also illustrates the image bounding boxes used to reconstruct problem images from publicly available KCSAT PDFs in the user environment, which makes the evaluation reproducible.

\begin{figure*}[t!]
  \centering
  \includegraphics[width=0.9\linewidth]{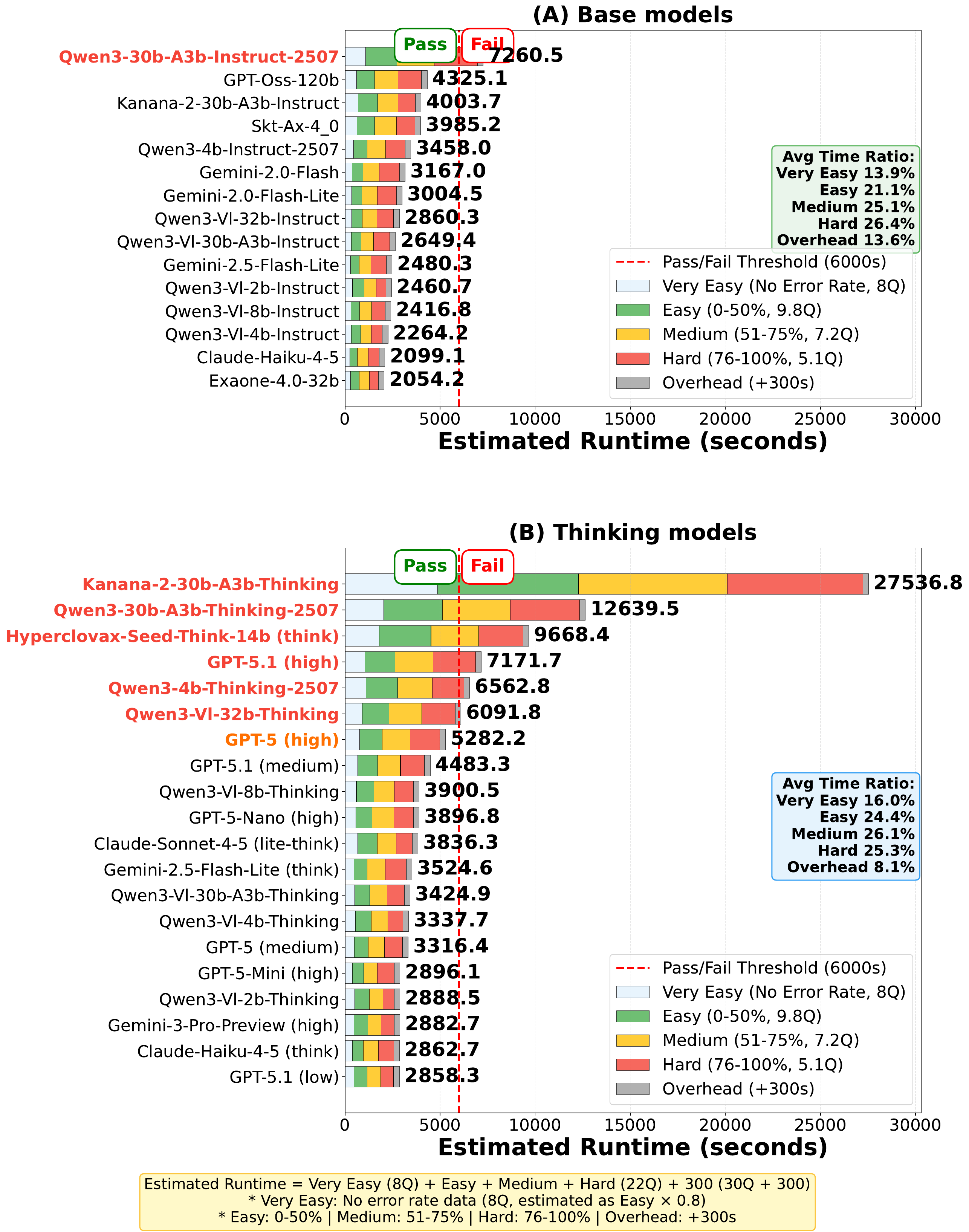}
    \caption{\textbf{Estimated wall-clock time to complete a full KCSAT math exam (6000\,s limit).} Per-tier average inference latencies weighted by expected question counts (Easy/Medium/Hard plus a ``Very Easy'' bin for items without published statistics), with a fixed pipeline overhead. (A) Base models; (B) Thinking models. Models left of the dashed line finish within time.}
  \label{fig:runtime_chart}
\end{figure*}

\paragraph{Runtime estimation under a 100-minute exam budget.}
Figure~\ref{fig:runtime_chart} addresses a practical question: if a model answers a full KCSAT mathematics form sequentially, would it finish within the official 100-minute limit (6000\,s)?
For each model, we estimate the total wall-clock time by summing tier-wise average inference latencies weighted by the expected number of questions in each tier. Tiers follow the Easy/Medium/Hard human error-rate bins defined in Sec.~\ref{sec:benchmark_construction}, augmented by a ``Very Easy'' bin that aggregates items without published cohort statistics (these items lie outside the 339-item core set and typically appear early in the form). A fixed system overhead for our evaluation pipeline (e.g., OCR/image handling and I/O) is then added.
Models whose estimated total time is below 6000\,s are marked as \textit{pass}.
This is an illustrative feasibility check under our protocol; absolute times may vary with hardware, batching, and API serving conditions.

\section{Detailed Numerical Results}
\label{sec:appendix_detail}

This appendix collects the per-model regression coefficients and DRG values referenced in Sections~\ref{sec:diff_signal}, \ref{sec:three_regimes}, and~\ref{sec:drg}.

\paragraph{Cohort error rate vs.\ examiner points (Sec.~\ref{sec:diff_signal}).}
Per-item mean model accuracy (averaged across all evaluated models) is regressed on each signal in turn. The cohort error rate yields $R^2 = 0.27$ (Pearson $r = -0.52$, $p{<}10^{-25}$) under wo.\,TTS, against $R^2 = 0.15$ for examiner-assigned points; under TTS the figures are $0.18$ and $0.05$. In a joint OLS regression with both signals as predictors, the standardised coefficient on cohort error rate is $\beta = -0.44$, twice that of examiner points ($\beta = -0.22$), so the cohort signal carries independent variance that the heuristic tier alone does not capture.

\paragraph{Per-model regression slopes/intercepts (Sec.~\ref{sec:three_regimes}).}
For each model with both wo.\,TTS and w.\,TTS runs, we regress per-item model error rate ($y$, in \%) on cohort error rate ($x$, in \%) separately for each condition. Top-row models in Figure~\ref{fig:three_regimes} (weak VLMs): HCX-Seed-Omni-8B (wo.\,TTS slope $0.25$, w.\,TTS slope $0.63$), Qwen3-VL-2B ($0.44 \to 0.88$), Qwen3-VL-4B ($0.63 \to 0.84$). Bottom-row models (frontier APIs): GPT-5.1 (wo.\,TTS slope $0.74$ / intercept $+\!12.6$, w.\,TTS slope $0.14$ / intercept $-9.4$), Claude-Opus-4.5 ($0.81 / +\!8.7 \to 0.32 / -9.4$). Gemini-3-Pro-Preview is a degenerate case under wo.\,TTS: \texttt{force-non-think} produces refusal-style outputs and yields mean accuracy $37\%$ with slope $0.18$; under w.\,TTS the model reaches $98\%$ accuracy with slope $0.01$ and intercept $-3.8$, placing it firmly in the below-cohort regime.

\paragraph{Monetary cost per correct answer (Sec.~\ref{sec:cost_accuracy}).}
Estimated API costs (USD) are computed from each provider's public per-token pricing. Among near-saturated closed-source systems, the cost per correct answer falls in a narrow band: GPT-5.1 \$0.064, Claude-Opus-4.5 \$0.073, Gemini-3-Pro-Preview \$0.081. Across closed-source models, several mid-tier configurations with larger reasoning budgets reach both higher accuracy and lower cost than higher-tier models under low budgets, so practical cost-effectiveness is governed more by budget allocation than by model tier.

\paragraph{Per-family DRG values (Sec.~\ref{sec:drg}).}
Across the 22 model families with both TTS-off and TTS-on runs, DRG values cluster broadly between $-7$ and $+30$ pp. Within the high-accuracy cluster (max-effort score in $[86, 99]$), five frontier families span a 22\,pp DRG range: GPT-5.1 ($+18.0$), Gemini-2.5-Flash ($+15.1$), GPT-5-mini ($-2.6$), Gemini-3-Flash-Preview ($-3.4$), and GPT-5 ($-4.0$). Same-family generations can split sharply too: Claude Sonnet-4 ($-4.0$) versus Sonnet-4.5 ($+24.3$). The DRG-by-budget correlation across multi-budget families is $r=+0.94$ between the lowest and highest TTS settings, so DRG saturates at the first TTS turn-on and barely shifts thereafter. Qualitative inspection of LOW-DRG and Easy-bin-failure models reveals systematic over-engineering: 2017 Math (Type-NA) \#5 (3-point, $24\%$ cohort error) draws a $\sim$$3{,}200$-token solution from HCX-Think-32B invoking the Lambert $W$ function, and a textbook geometric-series item triggers $\sim$$32{,}000$ tokens of thinking trace in GPT-OSS-120B before producing an incorrect answer.

\section{Robustness Analyses}
\label{sec:robustness}

Because the core set has 339 items and the Hard tier in particular contains 78 items, a natural concern is whether the difficulty-conditioned trends in Sec.~\ref{sec:results} could be driven by decoding noise or by the limited size of the high-difficulty subset. We address both factors here.

\paragraph{Per-model stability across repeated decoding.}
We re-ran a representative subset of 76 model$\times$budget configurations five times each on the 339 core items, keeping all other settings identical, and computed the per-configuration standard deviation of overall accuracy across runs. Table~\ref{tab:fiverun_stability} summarizes the distribution: the median standard deviation is $0.014$ and the mean is $0.037$ (in accuracy units), with $82.9\%$ of configurations exhibiting Std $<0.05$. The decoding noise on aggregated accuracy is therefore much smaller than the differences between difficulty tiers ($\geq 25$ percentage points) and between wo./w.~TTS settings ($\geq 30$ percentage points for most models) discussed in the main paper.

\begin{table}[t]
\centering
\small
\setlength{\tabcolsep}{6pt}
\begin{tabular}{lc}
\toprule
Statistic of per-config Std (5 runs) & Value \\
\midrule
N configurations & 76 \\
Mean Std (accuracy) & 0.037 \\
Median Std & 0.014 \\
Fraction with Std $<0.05$ & 82.9\% \\
\bottomrule
\end{tabular}
\caption{\textbf{Decoding stability across 5 repeated runs.} Per-configuration standard deviation of overall accuracy on the 339-item core set, across 76 model$\times$budget configurations. Noise is small relative to the difficulty- and budget-conditioned effects in Sec.~\ref{sec:results}.}
\label{tab:fiverun_stability}
\end{table}

\paragraph{Subset stability via $k$-fold splits on the Hard tier.}
To check whether the size of the Hard tier (78 items) is sufficient for the model-ranking conclusions drawn in Sec.~\ref{sec:diff_signal} and Figure~\ref{fig:model_vs_tts}, we partition the Hard tier into $k$ disjoint folds, compute per-fold model accuracy, and correlate the resulting per-fold ranking with the ranking obtained on the full Hard tier. Table~\ref{tab:kfold_stability} reports the Pearson correlation; even for $k{=}10$ (each fold contains only $7$--$8$ items), per-fold model rankings remain almost perfectly aligned with the full-Hard-tier ranking ($r = 0.975$). At $k{=}5$ ($15$--$16$ items per fold), correlation rises to $0.990$. This indicates that the relative ordering of models on Hard items is stable across random subsets and is not an artifact of the specific 78-item composition.

\begin{table}[t]
\centering
\small
\setlength{\tabcolsep}{6pt}
\begin{tabular}{ccc}
\toprule
$k$ (folds) & Items per fold & Pearson $r$ vs.\ full \\
\midrule
2  & 39     & 0.997 \\
3  & 26     & 0.996 \\
5  & 15--16 & 0.990 \\
10 & 7--8   & 0.975 \\
\bottomrule
\end{tabular}
\caption{\textbf{Subset stability of model rankings on the Hard tier.} For each $k$, the 78 Hard items are split into $k$ folds; per-fold rankings are correlated (Pearson) with the full-tier ranking and averaged. Rankings remain near-identical even for $k{=}10$.}
\label{tab:kfold_stability}
\end{table}

Together, these analyses indicate that (i) single-run aggregate accuracies are reliable proxies for repeated-run means, and (ii) the qualitative ranking of models on the Hard tier is robust to the specific subset of items used. The difficulty-conditioned conclusions in Sec.~\ref{sec:results} therefore do not hinge on the limited size of the 78-item Hard subset.

\section{Edge-Case Item Studies}
\label{sec:edge_cases}

Aggregate trends aside, KCSAT-ML's per-item human statistics let us isolate individual problems on which the difficulty-conditioned pattern is concretely visible. We pick two complementary edge cases from the 5-run logs of Sec.~\ref{sec:robustness}: an extreme-difficulty item where a $4\times$ smaller model with TTS clearly beats a larger sibling under low-budget decoding (anti-scaling), and an Easy item where the same strong model's TTS variant degrades the wo.\,TTS answer (overthinking).

\paragraph{Anti-scaling on extreme-difficulty items.}
On 2022 Mathematics Probability \& Statistics \#30 (cohort error rate $96.4\%$, the upper extreme of the Hard tier; reference answer $191$), Qwen3-VL-32B-Instruct under a strict low-budget setting (force-non-think) produces a $3$-token answer ``13'' and is wrong (\xmark); the same backbone never recovers under repeated trials in this configuration. In contrast, Qwen3-VL-8B-Thinking with TTS is correct in every one of the five repeated runs (\cmark), converging to $191$ via $\sim$$9{,}600$ structured reasoning tokens per trial. The smaller backbone with TTS therefore solves a problem on which a $4\times$ larger sibling under low-budget decoding fails completely.

The pattern is not item-specific. The upper block of Table~\ref{tab:edge_cases_combined} lists five Hard items (cohort error rates $93$--$98\%$) on which the anti-scaling outcome holds: a $4\times$ smaller w.\,TTS configuration solves the item near-perfectly while the larger wo.\,TTS configuration outputs a $\leq 4$-token answer and is wrong. The same effect appears against an even larger 30B mixture-of-experts baseline (Qwen3-30B-A3B-Instruct, force-non-think) on the identical item set, again pointing to inference-time compute rather than parameter count as the source of improvement.

\paragraph{Overthinking on Easy/Medium items.}
The converse pattern appears on items well within model competence under direct answering. The lower block of Table~\ref{tab:edge_cases_combined} lists five items spanning $32$--$58\%$ cohort error in which the \emph{same-backbone} Instruct (wo.\,TTS) variant solves the problem in every trial while the Thinking (w.\,TTS) variant fails the majority of trials despite spending $3$--$6{\times}$ more output tokens. The effect spans multiple families. For instance, on 2019 Mathematics (Type-NA) \#26 (cohort error rate $46.8\%$), Qwen3-VL-32B-Instruct is correct in all five runs (\cmark) with $\sim$$945$ output tokens per trial, whereas Qwen3-VL-32B-Thinking expends $\sim$$2{,}900$ tokens per trial and is wrong in every run (\xmark). As a more striking instance, on 2019 Mathematics (Type-GA) \#24 (cohort error rate $37.7\%$; reference answer $4$), GPT-OSS-120B under the low-budget setting answers ``4'' in $576$ output tokens (\cmark), but enabling TTS on the same backbone produces $3{,}000$--$4{,}000$-token derivations that miscalculate to $\boxed{4\sqrt{13}}$ or $\boxed{16}$ in all five trials (\xmark).

\begin{table*}[t]
\centering
\small
\setlength{\tabcolsep}{4pt}
\begin{adjustbox}{max width=\textwidth}
\begin{tabular}{lcll}
\toprule
\textbf{Item (KCSAT)} & \textbf{Cohort err.} & \textbf{wo.\,TTS (maj.\ / avg tok)} & \textbf{w.\,TTS (maj.\ / avg tok)} \\
\midrule
\multicolumn{4}{l}{\textit{(a) Anti-scaling: Qwen3-VL-32B-Instruct (force-non-think) wo.\,TTS  vs.  Qwen3-VL-8B-Thinking w.\,TTS}} \\
\midrule
2022 Prob.\,\&\,Stat.\ \#30   & 96.4\% & \xmark{} \,/\, 3 tok & \cmark{} \,/\, $\sim$9{,}600 tok  \\
2020 Math NA \#30             & 97.9\% & \xmark{} \,/\, 2 tok & \cmark{} \,/\, $\sim$11{,}400 tok \\
2018 Math NA \#30             & 93.4\% & \xmark{} \,/\, 3 tok & \cmark{} \,/\, $\sim$15{,}800 tok \\
2020 Math GA \#30             & 95.5\% & \xmark{} \,/\, 3 tok & \cmark{} \,/\, $\sim$5{,}500 tok  \\
2024 Math Calc.\ \#29         & 93.2\% & \xmark{} \,/\, 3 tok & \cmark{} \,/\, $\sim$10{,}900 tok \\
\midrule
\multicolumn{4}{l}{\textit{(b) Overthinking: same-backbone Instruct wo.\,TTS  vs.  Thinking w.\,TTS}} \\
\midrule
2019 Math NA \#26 \emph{[Qwen3-VL-32B]}    & 46.8\% & \cmark{} \,/\, 945 tok           & \xmark{} \,/\, $\sim$2{,}900 tok  \\
2025 Math Calc.\ \#26 \emph{[Qwen3-VL-32B]} & 32.2\% & \cmark{} \,/\, $\sim$1{,}000 tok & \xmark{} \,/\, $\sim$3{,}400 tok \\
2024 Math Geo.\ \#27 \emph{[Qwen3-VL-32B]}  & 58.2\% & \cmark{} \,/\, $\sim$3{,}600 tok & \xmark{} \,/\, $\sim$14{,}800 tok \\
2014 Math A \#18 \emph{[Qwen3-30B-A3B]}     & 43.8\% & \cmark{} \,/\, 718 tok           & \xmark{} \,/\, $\sim$2{,}300 tok  \\
2019 Math GA \#24 \emph{[GPT-OSS-120B]}     & 37.7\% & \cmark{} \,/\, 576 tok           & \xmark{} \,/\, $\sim$3{,}400 tok  \\
\bottomrule
\end{tabular}
\end{adjustbox}
\caption{\textbf{Edge-case items showing two faces of difficulty-conditional TTS.} Cells report the majority outcome across repeated trials (\cmark{} = correct majority, \xmark{} otherwise) and average output tokens. (a) Anti-scaling: a smaller w.\,TTS variant solves Hard items that a larger wo.\,TTS baseline fails. (b) Overthinking: on easier items, the same-backbone Thinking variant fails despite spending more tokens. Model family varies per row in (b), shown in italics.}
\label{tab:edge_cases_combined}
\end{table*}

Together, blocks (a) and (b) show that the same model family can shift from anti-scaling (small+TTS $>$ big wo.\,TTS) to overthinking (wo.\,TTS $>$ w.\,TTS) within one benchmark, purely as a function of item difficulty.

\begin{figure*}[t!]
  \centering
  \footnotesize
  \begin{tcolorbox}[enhanced,width=0.95\linewidth,colback=white,colframe=bluee!55,
      arc=1mm,boxrule=0.6pt,left=8pt,right=8pt,top=6pt,bottom=6pt,
      title={\textbf{KCSAT-ML geometry item} \,(2025 KCSAT, \#28, 4 points)},
      coltitle=white,colbacktitle=bluee!75,fonttitle=\bfseries\small,
      attach boxed title to top left={xshift=6pt,yshift=-2mm},
      boxed title style={colframe=bluee!75,arc=0.5mm}]
    \begin{tcolorbox}[enhanced,width=\linewidth,colback=white,colframe=black!30,
        boxrule=0.5pt,arc=0.6mm,drop shadow={black!35,opacity=0.45,xshift=1pt,yshift=-1pt},
        left=11pt,right=9pt,top=8pt,bottom=9pt,before skip=2pt,after skip=4pt]
      \begin{minipage}[c]{0.575\linewidth}
        {\setlength{\parindent}{0pt}\hangindent=1.7em\hangafter=1\relax
         {\bfseries 28.}\hspace{0.5em}In coordinate space, right triangle $ABC$ has $\overline{AB}=8$, $\overline{BC}=6$, and $\angle ABC=\tfrac{\pi}{2}$. Let $S$ be the sphere whose diameter is $\overline{AC}$. The plane that contains line $AB$ and is perpendicular to plane $ABC$ meets $S$ in a circle $O$. Let $P,Q$ be the two distinct points of $O$ whose distance from line $AC$ equals $4$. Find $\overline{PQ}$.\;{\footnotesize\color{black!55}[4 points]}\par}
        \vspace{7pt}
        \mbox{\ding{172}\,$\sqrt{43}$}\quad\mbox{\ding{173}\,$\sqrt{47}$}\quad\mbox{\ding{174}\,$\sqrt{51}$}\quad\mbox{\ding{175}\,$\sqrt{55}$}\quad\mbox{\ding{176}\,$\sqrt{59}$}
      \end{minipage}\hfill
      \begin{minipage}[c]{0.39\linewidth}
        \centering
        \begin{tcolorbox}[enhanced,width=\linewidth,height=3.3cm,halign=center,valign=center,
            colback=black!16,colframe=black!35,boxrule=0.5pt,arc=2pt,
            left=2pt,right=2pt,top=2pt,bottom=2pt]
          \begin{tikzpicture}
            \draw[black!55,line width=0.8pt,rounded corners=1.2pt] (-0.62,0.20) rectangle (0.62,1.02);
            \fill[black!55] (0.32,0.80) circle (0.085);
            \draw[black!55,line width=0.9pt,line join=round]
                  (-0.59,0.42) -- (-0.18,0.72) -- (0.08,0.54) -- (0.59,0.97);
            \node[text=black!60,font=\itshape] at (0,-0.16) {Figure};
          \end{tikzpicture}
        \end{tcolorbox}
      \end{minipage}
    \end{tcolorbox}
    {\scriptsize Ground truth: \ding{175}\ $\sqrt{55}$.}

    \medskip
    \tcbline
    \textbf{Response model:}\; \texttt{Gemini-2.5-Flash}

    \smallskip
    \begin{tcolorbox}[enhanced,colback=redd!6,colframe=redd!45,arc=0.6mm,boxrule=0.4pt,
        left=5pt,right=5pt,top=3pt,bottom=3pt,
        title={\textbf{wo.\,TTS}\; \footnotesize\textit{Prompt:} ``Solve the problem. Provide ONLY the final answer; no reasoning.''},
        coltitle=black,colbacktitle=redd!18,fonttitle=\small]
      \textbf{Response:}\; $2\sqrt{7}$ \quad \textcolor{redd}{\xmark\ \textit{incorrect}}
    \end{tcolorbox}

    \smallskip
    \begin{tcolorbox}[enhanced,colback=bluee!6,colframe=bluee!45,arc=0.6mm,boxrule=0.4pt,
        left=5pt,right=5pt,top=3pt,bottom=3pt,
        title={\textbf{w.\,TTS}\; \footnotesize\textit{Prompt:} ``Solve the problem.''},
        coltitle=black,colbacktitle=bluee!18,fonttitle=\small]
      \textbf{Response} (abridged chain-of-thought):\;
      Set $B=(0,0,0)$, $A=(8,0,0)$, $C=(0,6,0)$, so plane $ABC$ is $z=0$ and $\overline{AC}=10$.
      Sphere $S$ has center $=$ midpoint of $\overline{AC}$ and radius $5$. The plane through $AB$
      perpendicular to $ABC$ is $y=0$; its intersection with $S$ is circle $O$. Imposing the
      distance-$4$-from-line-$AC$ condition on $O$ locates $P,Q$, giving
      $\overline{PQ}=\boxed{\sqrt{55}}$.\quad \textcolor{greene}{\cmark\ \textit{correct}}
    \end{tcolorbox}
  \end{tcolorbox}
  \caption{\textbf{Example of a geometry-based KCSAT-ML item.} A representative geometry problem (2025, \#28) with sample wo.\,TTS and w.\,TTS responses from Gemini-2.5-Flash. The accompanying diagram (right) is omitted in this example and shown only as a placeholder.}
  \label{fig:geometry_example}
\end{figure*}

\begin{figure*}[t!]
  \centering
  \footnotesize
  \begin{tcolorbox}[enhanced,width=0.95\linewidth,colback=white,colframe=bluee!55,
      arc=1mm,boxrule=0.6pt,left=8pt,right=8pt,top=6pt,bottom=6pt,
      title={\textbf{KCSAT-ML algebra item} \,(2025 KCSAT, Common \#6, 3 points)},
      coltitle=white,colbacktitle=bluee!75,fonttitle=\bfseries\small,
      attach boxed title to top left={xshift=6pt,yshift=-2mm},
      boxed title style={colframe=bluee!75,arc=0.5mm}]
    \begin{tcolorbox}[enhanced,width=\linewidth,colback=white,colframe=black!30,
        boxrule=0.5pt,arc=0.6mm,drop shadow={black!35,opacity=0.45,xshift=1pt,yshift=-1pt},
        left=11pt,right=9pt,top=8pt,bottom=9pt,before skip=2pt,after skip=4pt]
      {\setlength{\parindent}{0pt}\hangindent=1.75em\hangafter=1\relax
       {\bfseries 6.}\hspace{0.55em}Given $\cos\!\left(\dfrac{\pi}{2}+\theta\right)=-\dfrac{1}{5}$,\;
       find the value of $\dfrac{\sin\theta}{\,1-\cos^{2}\theta\,}$.\;{\footnotesize\color{black!55}[3 points]}\par}
      \vspace{7pt}
      \noindent\makebox[\linewidth]{\hspace*{1em}%
        \ding{172}\,$-5$\hfill\ding{173}\,$-\sqrt{5}$\hfill\ding{174}\,$0$%
        \hfill\ding{175}\,$\sqrt{5}$\hfill\ding{176}\,$5$\hspace*{0.6em}}
    \end{tcolorbox}
    {\scriptsize Ground truth: \ding{176}\ $5$.}

    \medskip
    \tcbline
    \textbf{Response model:}\; \texttt{HyperCLOVAX-SEED-Think-14B}

    \smallskip
    \begin{tcolorbox}[enhanced,colback=redd!6,colframe=redd!45,arc=0.6mm,boxrule=0.4pt,
        left=5pt,right=5pt,top=3pt,bottom=3pt,
        title={\textbf{wo.\,TTS}\; \footnotesize\textit{Prompt:} ``Solve the problem. Provide ONLY the final answer; no reasoning.''},
        coltitle=black,colbacktitle=redd!18,fonttitle=\small]
      \textbf{Response:}\; $5$ \quad \textcolor{greene}{\cmark\ \textit{correct}}
    \end{tcolorbox}

    \smallskip
    \begin{tcolorbox}[enhanced,colback=bluee!6,colframe=bluee!45,arc=0.6mm,boxrule=0.4pt,
        left=5pt,right=5pt,top=3pt,bottom=3pt,
        title={\textbf{w.\,TTS}\; \footnotesize\textit{Prompt:} ``Solve the problem.''},
        coltitle=black,colbacktitle=bluee!18,fonttitle=\small]
      \textbf{Response} (abridged chain-of-thought):\;
      By the co-function identity, $\cos\!\left(\tfrac{\pi}{2}+\theta\right)=-\sin\theta$,
      so $-\sin\theta=-\tfrac{1}{5}\Rightarrow \sin\theta=\tfrac{1}{5}$.
      Since $1-\cos^{2}\theta=\sin^{2}\theta$, we obtain
      $\dfrac{\sin\theta}{1-\cos^{2}\theta}=\dfrac{\sin\theta}{\sin^{2}\theta}
      =\dfrac{1}{\sin\theta}=5$.
      Final answer: $\boxed{5}$ \quad \textcolor{greene}{\cmark\ \textit{correct}}
    \end{tcolorbox}
  \end{tcolorbox}
  \caption{\textbf{Example of an algebraic KCSAT-ML item.} A representative algebra problem (2025, \#6) with sample wo.\,TTS and w.\,TTS responses from HyperCLOVAX-SEED-Think-14B.}
  \label{fig:kcsat_algebra_example}
\end{figure*}

\begin{table}[t]
\centering
\begin{adjustbox}{width=\columnwidth}
\begin{tabular}{lcc}
\toprule
\textbf{Statistic} & \textbf{KCSAT-ML (Full)} & \textbf{KCSAT-ML (Core)} \\
\midrule
Images & 664 & 339 \\
Questions & 664 & 339 \\
\midrule
\textit{Question Type} & & \\
\quad Multiple-choice & 468 (70.5\%) & -- \\
\quad Short-answer & 196 (29.5\%) & 339 (100.0\%) \\
\midrule
Subjects & Mathematics (100\%) & Mathematics (100\%) \\
\midrule
\textit{Answer Type} & & \\
\quad Integer & 450 (67.8\%) & 227 (67.0\%) \\
\quad Fraction & 142 (21.4\%) & 66 (19.5\%) \\
\quad Expression & 37 (5.6\%) & 18 (5.3\%) \\
\quad Options & 18 (2.7\%) & 17 (5.0\%) \\
\quad Decimal & 17 (2.6\%) & 11 (3.2\%) \\
\midrule
\textit{Points} & & \\
\quad 2 points & 68 (10.2\%) & 0 (0.0\%) \\
\quad 3 points & 313 (47.1\%) & 76 (22.4\%) \\
\quad 4 points & 283 (42.6\%) & 263 (77.6\%) \\
\midrule
\textit{Human Error Rate} & & \\
\quad None & 325 (48.9\%) & 0 (0.0\%) \\
\quad Easy (0--50\%) & 148 (22.3\%) & 148 (43.7\%) \\
\quad Medium (51--75\%) & 113 (17.0\%) & 113 (33.3\%) \\
\quad Hard (76--100\%) & 78 (11.7\%) & 78 (23.0\%) \\
\bottomrule
\end{tabular}
\end{adjustbox}
\caption{\label{tab:kcsatml_statistics}
\textbf{Statistics of KCSAT-ML.} Full benchmark (664 items) and the core subset (339 short-answer items with official human error rates) used in all difficulty-conditioned analyses.}
\end{table}

\begin{table*}[t]
\centering
\setlength{\tabcolsep}{2pt}

\begin{tabular}{@{}p{0.49\textwidth}@{\hspace{0.02\textwidth}}p{0.49\textwidth}@{}}

\begin{minipage}[t]{\linewidth}
\centering
\caption*{\textbf{Open-Weight Models}}
\begin{adjustbox}{width=\linewidth}
\begin{tabular}{lccc}
\toprule
\textbf{Model} & \textbf{TTS} & \textbf{Avg.Tok} & \textbf{Score} \\
\midrule
\multicolumn{4}{l}{\textit{\textbf{Open-Weight LLMs (Korean \& Global)}}} \\
\midrule
\rowcolor{gray!10} Solar-10-7B-Instruct-v1.0 & & 85 & 3.2 \\
\rowcolor{bluee!20} \textbf{Solar-10-7B-Instruct-v1.0} & \checkmark & \textbf{700} & \textbf{0.0} \\
\addlinespace[0.1cm]
\rowcolor{gray!10} Solar-Pro-Preview-Instruct-22B & & 5 & 2.4 \\
\rowcolor{bluee!20} \textbf{Solar-Pro-Preview-Instruct-22B} & \checkmark & \textbf{686} & \textbf{0.0} \\
\addlinespace[0.1cm]
\rowcolor{gray!10} Midm-2.0-Base-Instruct-12B & & 11 & 3.8 \\
\rowcolor{bluee!20} \textbf{Midm-2.0-Base-Instruct-12B} & \checkmark & \textbf{695} & \textbf{8.6} \\
\addlinespace[0.1cm]
\rowcolor{gray!10} SKT-AX-4.0-Light-7B & & 465 & 22.7 \\
\rowcolor{bluee!20} \textbf{SKT-AX-4.0-Light-7B} & \checkmark & \textbf{1,002} & \textbf{14.2} \\
\addlinespace[0.1cm]
\rowcolor{gray!10} SKT-AX-4.0-72B & & 9 & 5.3 \\
\rowcolor{bluee!20} \textbf{SKT-AX-4.0-72B} & \checkmark & \textbf{1,873} & \textbf{20.1} \\
\addlinespace[0.1cm]
\rowcolor{gray!10} GPT-OSS-20B & & 2,272 & 67.8 \\
\rowcolor{bluee!20} \textbf{GPT-OSS-20B} & \checkmark & \textbf{4,283} & \textbf{58.4} \\
\addlinespace[0.1cm]
\rowcolor{gray!10} Exaone-4.0-32B & & 6 & 3.8 \\
\rowcolor{bluee!20} \textbf{Exaone-4.0-32B} & \checkmark & \textbf{2,396} & \textbf{39.8} \\
\addlinespace[0.1cm]
\rowcolor{gray!10} Kanana-2-30B-A3B-Instruct & & 9 & 5.6 \\
\rowcolor{bluee!10} \textbf{Kanana-2-30B-A3B-Instruct} & \checkmark & \textbf{1,324} & \textbf{27.1} \\
\rowcolor{bluee!10} \textbf{Kanana-2-30B-A3B-Thinking} & \checkmark & \textbf{9,875} & \textbf{45.4} \\
\addlinespace[0.1cm]
\rowcolor{gray!10} HyperCLOVA-X-Seed-14B & & 55 & 6.8 \\
\rowcolor{bluee!10} \textbf{HyperCLOVA-X-Seed-14B(Non-think)} & \checkmark & \textbf{990} & \textbf{8.8} \\
\rowcolor{bluee!20} \textbf{HyperCLOVA-X-Seed-14B(Think)} & \checkmark & \textbf{8,646} & \textbf{53.7} \\
\addlinespace[0.1cm]
\rowcolor{gray!10} Qwen3-4B-Instruct & & 7 & 5.9 \\
\rowcolor{bluee!10} \textbf{Qwen3-4B-Instruct} & \checkmark & \textbf{4,274} & \textbf{49.6} \\
\rowcolor{bluee!20} \textbf{Qwen3-4B-Thinking} & \checkmark & \textbf{8,558} & \textbf{64.3} \\
\addlinespace[0.1cm]
\rowcolor{gray!10} GPT-OSS-120B & & 1,839 & 74.9 \\
\rowcolor{bluee!10} \textbf{GPT-OSS-120B} & \checkmark & \textbf{3,552} & \textbf{67.6} \\
\addlinespace[0.1cm]
\rowcolor{gray!10} Qwen3-30B-A3B-Instruct & & 5 & 7.4 \\
\rowcolor{bluee!10} \textbf{Qwen3-30B-A3B-Instruct} & \checkmark & \textbf{4,063} & \textbf{62.2} \\
\rowcolor{bluee!20} \textbf{Qwen3-30B-A3B-Thinking} & \checkmark & \textbf{7,678} & \textbf{74.0} \\
\midrule
\rowcolor{white} \multicolumn{4}{l}{\textit{\textbf{Open-Weight VLMs}}} \\
\midrule
\rowcolor{gray!10} HyperCLOVA-X-Vision-3B & & 9 & 1.5 \\
\rowcolor{bluee!20} \textbf{HyperCLOVA-X-Vision-3B} & \checkmark & \textbf{749} & \textbf{4.4} \\
\addlinespace[0.1cm]
\rowcolor{gray!10} HyperCLOVA-X-Seed-Think-32B(Non-think) & & 746 & 39.5 \\
\rowcolor{bluee!20} \textbf{HyperCLOVA-X-Seed-Think-32B(Think)} & \checkmark & \textbf{7,431} & \textbf{87.2} \\
\addlinespace[0.1cm]
\rowcolor{gray!10} Qwen3-VL-2B-Instruct & & 821 & 1.2 \\
\rowcolor{bluee!10} \textbf{Qwen3-VL-2B-Instruct} & \checkmark & \textbf{10,650} & \textbf{5.9} \\
\rowcolor{bluee!20} \textbf{Qwen3-VL-2B-Thinking} & \checkmark & \textbf{12,379} & \textbf{18.3} \\
\addlinespace[0.1cm]
\rowcolor{gray!10} Qwen3-VL-4B-Instruct & & 112 & 3.2 \\
\rowcolor{bluee!10} \textbf{Qwen3-VL-4B-Instruct} & \checkmark & \textbf{6,668} & \textbf{28.3} \\
\rowcolor{bluee!20} \textbf{Qwen3-VL-4B-Thinking} & \checkmark & \textbf{9,975} & \textbf{35.4} \\
\addlinespace[0.1cm]
\rowcolor{gray!10} Qwen3-VL-30B-A3B-Instruct & & 92 & 7.7 \\
\rowcolor{bluee!10} \textbf{Qwen3-VL-30B-A3B-Instruct} & \checkmark & \textbf{5,699} & \textbf{35.4} \\
\rowcolor{bluee!20} \textbf{Qwen3-VL-30B-A3B-Thinking} & \checkmark & \textbf{7,602} & \textbf{45.4} \\
\addlinespace[0.1cm]
\rowcolor{gray!10} Qwen3-VL-8B-Instruct & & 58 & 6.5 \\
\rowcolor{bluee!10} \textbf{Qwen3-VL-8B-Instruct} & \checkmark & \textbf{5,450} & \textbf{46.3} \\
\rowcolor{bluee!20} \textbf{Qwen3-VL-8B-Thinking} & \checkmark & \textbf{8,842} & \textbf{52.2} \\
\addlinespace[0.1cm]
\rowcolor{gray!10} Qwen3-VL-32B-Instruct & & 5 & 6.5 \\
\rowcolor{bluee!10} \textbf{Qwen3-VL-32B-Instruct} & \checkmark & \textbf{3,703} & \textbf{54.6} \\
\rowcolor{bluee!20} \textbf{Qwen3-VL-32B-Thinking} & \checkmark & \textbf{7,808} & \textbf{61.7} \\
\bottomrule
\end{tabular}
\end{adjustbox}
\end{minipage}

&

\begin{minipage}[t]{\linewidth}
\centering
\caption*{\textbf{Closed-Source APIs}}
\begin{adjustbox}{width=0.9\linewidth}
\begin{tabular}{lccc}
\toprule
\textbf{Model} & \textbf{TTS} & \textbf{Avg.Tok} & \textbf{Score} \\
\midrule
\rowcolor{gray!10} GPT-5-Nano (Minimal) & & 10 & 1.2 \\
\rowcolor{bluee!10} \textbf{GPT-5-Nano (Minimal)} & \checkmark & \textbf{657} & \textbf{2.7} \\
\rowcolor{bluee!10} \textbf{GPT-5-Nano (low)} & \checkmark & \textbf{1,697} & \textbf{29.8} \\
\rowcolor{bluee!10} \textbf{GPT-5-Nano (Medium)} & \checkmark & \textbf{6,446} & \textbf{54.0} \\
\rowcolor{bluee!20} \textbf{GPT-5-Nano (High)} & \checkmark & \textbf{13,079} & \textbf{60.8} \\
\addlinespace[0.1cm]
\rowcolor{gray!10} o4-Mini (Low) & & 9 & 73.2 \\
\rowcolor{bluee!10} \textbf{o4-Mini (Low)} & \checkmark & \textbf{1,674} & \textbf{59.9} \\
\rowcolor{bluee!10} \textbf{o4-Mini (Medium)} & \checkmark & \textbf{2,976} & \textbf{74.3} \\
\rowcolor{bluee!20} \textbf{o4-Mini (High)} & \checkmark & \textbf{5,350} & \textbf{83.5} \\
\addlinespace[0.1cm]
\rowcolor{gray!10} GPT-5-Mini (Minimal) & & 9 & 5.0 \\
\rowcolor{bluee!10} \textbf{GPT-5-Mini (Minimal)} & \checkmark & \textbf{957} & \textbf{21.8} \\
\rowcolor{bluee!10} \textbf{GPT-5-Mini (Low)} & \checkmark & \textbf{1,495} & \textbf{62.2} \\
\rowcolor{bluee!10} \textbf{GPT-5-Mini (Medium)} & \checkmark & \textbf{2,741} & \textbf{76.7} \\
\rowcolor{bluee!20} \textbf{GPT-5-Mini (High)} & \checkmark & \textbf{5,104} & \textbf{86.1} \\
\addlinespace[0.1cm]
\rowcolor{gray!10} GPT-5 (Minimal) & & 8 & 7.7 \\
\rowcolor{bluee!10} \textbf{GPT-5 (Minimal)} & \checkmark & \textbf{538} & \textbf{17.1} \\
\rowcolor{bluee!10} \textbf{GPT-5 (Low)} & \checkmark & \textbf{2,000} & \textbf{82.9} \\
\rowcolor{bluee!10} \textbf{GPT-5 (Medium)} & \checkmark & \textbf{3,808} & \textbf{91.2} \\
\rowcolor{bluee!20} \textbf{GPT-5 (High)} & \checkmark & \textbf{6,442} & \textbf{94.1} \\
\addlinespace[0.1cm]
\rowcolor{gray!10} GPT-5.1 (None) & & 13 & 8.8 \\
\rowcolor{bluee!10} \textbf{GPT-5.1 (None)} & \checkmark & \textbf{691} & \textbf{15.0} \\
\rowcolor{bluee!10} \textbf{GPT-5.1 (Low)} & \checkmark & \textbf{1,916} & \textbf{73.7} \\
\rowcolor{bluee!10} \textbf{GPT-5.1 (Medium)} & \checkmark & \textbf{3,933} & \textbf{90.0} \\
\rowcolor{bluee!20} \textbf{GPT-5.1 (High)} & \checkmark & \textbf{6,587} & \textbf{93.2} \\
\addlinespace[0.1cm]
\rowcolor{gray!10} Claude-Sonnet-4 (Non-think) & & 328 & 15.6 \\
\rowcolor{bluee!10} \textbf{Claude-Sonnet-4} & \checkmark & \textbf{827} & \textbf{16.5} \\
\rowcolor{bluee!10} \textbf{Claude-Sonnet-4 (Lite-think)} & \checkmark & \textbf{2,201} & \textbf{23.9} \\
\rowcolor{bluee!20} \textbf{Claude-Sonnet-4 (Think)} & \checkmark & \textbf{3,594} & \textbf{31.9} \\
\addlinespace[0.1cm]
\rowcolor{gray!10} Claude-Sonnet-4.5 (Non-think) & & 549 & 33.3 \\
\rowcolor{bluee!10} \textbf{Claude-Sonnet-4.5} & \checkmark & \textbf{868} & \textbf{19.8} \\
\rowcolor{bluee!10} \textbf{Claude-Sonnet-4.5 (Lite-think)} & \checkmark & \textbf{2,524} & \textbf{41.6} \\
\rowcolor{bluee!20} \textbf{Claude-Sonnet-4.5 (Think)} & \checkmark & \textbf{3,732} & \textbf{49.3} \\
\addlinespace[0.1cm]
\rowcolor{gray!10} Claude-Haiku-4.5 (Non-think) & & 66 & 7.7 \\
\rowcolor{bluee!10} \textbf{Claude-Haiku-4.5} & \checkmark & \textbf{833} & \textbf{13.9} \\
\rowcolor{bluee!10} \textbf{Claude-Haiku-4.5 (Lite-think)} & \checkmark & \textbf{2,824} & \textbf{25.7} \\
\rowcolor{bluee!20} \textbf{Claude-Haiku-4.5 (Think)} & \checkmark & \textbf{4,440} & \textbf{33.3} \\
\addlinespace[0.1cm]
\rowcolor{gray!10} Claude-Opus-4.5 (Non-think) & & 7 & 13.3 \\
\rowcolor{bluee!10} \textbf{Claude-Opus-4.5} & \checkmark & \textbf{736} & \textbf{44.5} \\
\rowcolor{bluee!10} \textbf{Claude-Opus-4.5 (Lite-think)} & \checkmark & \textbf{2,477} & \textbf{79.1} \\
\rowcolor{bluee!20} \textbf{Claude-Opus-4.5 (Think)} & \checkmark & \textbf{3,617} & \textbf{87.6} \\
\addlinespace[0.1cm]
\rowcolor{gray!10} Gemini-2.5-Flash-Lite & & 3 & 6.9 \\
\rowcolor{bluee!10} \textbf{Gemini-2.5-Flash-Lite} & \checkmark & \textbf{4,850} & \textbf{8.8} \\
\rowcolor{bluee!20} \textbf{Gemini-2.5-Flash-Lite (Think)} & \checkmark & \textbf{11,933} & \textbf{66.7} \\
\rowcolor{gray!10} Gemini-2.5-Flash (Non-think) & & 4 & 11.1 \\
\rowcolor{bluee!10} \textbf{Gemini-2.5-Flash} & \checkmark & \textbf{3,218} & \textbf{17.7} \\
\rowcolor{bluee!20} \textbf{Gemini-2.5-Flash (Think)} & \checkmark & \textbf{7,595} & \textbf{86.7} \\
\addlinespace[0.1cm]
\rowcolor{gray!10} Gemini-2.5-Pro (Think) & & 3 & 96.1 \\
\rowcolor{bluee!20} \textbf{Gemini-2.5-Pro (Think)} & \checkmark & \textbf{6,886} & \textbf{93.2} \\
\addlinespace[0.1cm]
\rowcolor{gray!10} Gemini-3-Flash-Prev (Minimal) & & 1,028 & 76.3 \\
\rowcolor{bluee!10} \textbf{Gemini-3-Flash-Prev (Minimal)} & \checkmark & \textbf{1,008} & \textbf{62.5} \\
\rowcolor{bluee!10} \textbf{Gemini-3-Flash-Prev (Low)} & \checkmark & \textbf{3,317} & \textbf{88.2} \\
\rowcolor{bluee!10} \textbf{Gemini-3-Flash-Prev (Medium)} & \checkmark & \textbf{7,362} & \textbf{96.2} \\
\rowcolor{bluee!20} \textbf{Gemini-3-Flash-Prev (High)} & \checkmark & \textbf{6,210} & \textbf{98.5} \\
\addlinespace[0.1cm]
\rowcolor{gray!10} Gemini-3-Pro-Prev (Low) & & 23 & 37.4 \\
\rowcolor{bluee!10} \textbf{Gemini-3-Pro-Prev (Low)} & \checkmark & \textbf{2,345} & \textbf{83.8} \\
\rowcolor{bluee!20} \textbf{Gemini-3-Pro-Prev (High)} & \checkmark & \textbf{6,913} & \textbf{99.4} \\
\bottomrule
\end{tabular}
\end{adjustbox}
\end{minipage}
\end{tabular}
\caption{\textbf{Full results on KCSAT-ML.} Complete results across all models and settings; extends Table~\ref{tab:main_results}. Baselines are shaded gray, their TTS counterparts blue.}
\label{tab:full_results}
\end{table*}

\begin{figure*}[t!]
  \centering
  \includegraphics[width=0.9\linewidth]{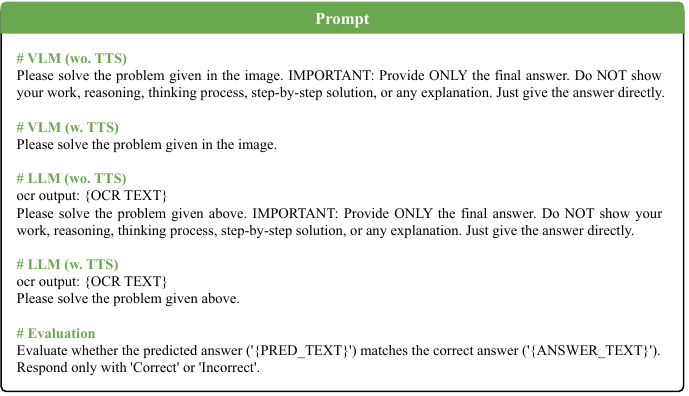}
    \caption{\textbf{Prompt templates used in experiments.} VLM and LLM settings under wo.\,TTS and w.\,TTS, plus the equivalence-judge prompt for answer verification.}
  \label{fig:prompt}
\end{figure*}

\begin{figure*}[t]
\centering
\begin{minipage}[c]{0.40\linewidth}
\centering
\begin{tikzpicture}[
    scale=1.0,
    every node/.style={transform shape},
    gline/.style={draw=black!28, line width=1.1pt, line cap=round},
    bubble/.style={draw=black!45, line width=0.5pt, fill=white, circle, minimum size=4pt, inner sep=0pt},
    numbadge/.style={draw=black!70, line width=0.6pt, fill=black!6, rounded corners=1pt, inner sep=1.5pt, font=\bfseries\scriptsize},
  ]
  \fill[black!12, rounded corners=3pt] (0.10,-0.10) rectangle (6.10,8.40);
  \draw[fill=white, draw=black!35, line width=0.6pt, rounded corners=3pt] (0,0) rectangle (6.0,8.5);

  \fill[bluee!18, rounded corners=2pt] (0.25,7.85) rectangle (5.75,8.30);
  \node[anchor=west, font=\footnotesize\bfseries, text=bluee!75!black] at (0.40,8.075) {Mathematics};
  \node[draw=bluee!70!black, fill=white, rounded corners=2pt, line width=0.5pt,
        font=\scriptsize\bfseries, text=bluee!70!black, inner sep=1.5pt] at (5.25,8.075) {p.19};

  \draw[black!20, line width=0.5pt] (3.0,0.35) -- (3.0,7.55);

  \node[numbadge] at (0.55,7.40) {27};
  \draw[gline] (0.95,7.42) -- (2.75,7.42);
  \draw[gline] (0.30,7.05) -- (2.75,7.05);
  \draw[gline] (0.30,6.78) -- (2.45,6.78);
  \foreach \i in {0,1,2,3,4}{
    \node[bubble] at (0.45+\i*0.52,6.35) {};
    \node[font=\tiny, text=black!55] at (0.45+\i*0.52,6.35) {\the\numexpr\i+1\relax};
  }

  \node[numbadge] at (0.55,5.85) {28};
  \draw[gline] (0.95,5.87) -- (2.75,5.87);
  \draw[gline] (0.30,5.52) -- (2.75,5.52);
  \draw[gline] (0.30,5.25) -- (2.30,5.25);
  \draw[black!55, line width=0.6pt] (0.45,4.55) -- (1.15,4.55) -- (0.80,5.05) -- cycle;
  \draw[gline] (1.45,4.78) -- (2.75,4.78);
  \draw[gline] (0.30,4.45) -- (2.75,4.45);
  \foreach \i in {0,1,2,3,4}{
    \node[bubble] at (0.45+\i*0.52,3.95) {};
    \node[font=\tiny, text=black!55] at (0.45+\i*0.52,3.95) {\the\numexpr\i+1\relax};
  }

  \node[numbadge] at (0.55,3.40) {29};
  \draw[gline] (0.95,3.42) -- (2.75,3.42);
  \draw[gline] (0.30,3.07) -- (2.75,3.07);
  \draw[gline] (0.30,2.80) -- (2.55,2.80);
  \draw[gline] (0.30,2.53) -- (2.75,2.53);
  \draw[gline] (0.30,2.26) -- (2.10,2.26);

  \node[numbadge] at (3.30,7.40) {28};
  \draw[gline] (3.70,7.42) -- (5.55,7.42);
  \draw[gline] (3.10,7.05) -- (5.55,7.05);
  \draw[black!55, line width=0.6pt, fill=black!4] (4.05,6.55) circle (0.32);
  \draw[black!45, line width=0.5pt] (3.73,6.55) arc (180:360:0.32 and 0.11);
  \draw[black!45, line width=0.5pt, densely dashed] (3.73,6.55) arc (180:360:0.32 and -0.11);
  \draw[gline] (4.55,6.62) -- (5.55,6.62);
  \draw[gline] (3.10,6.20) -- (5.55,6.20);
  \draw[gline] (3.10,5.93) -- (5.10,5.93);
  \foreach \i in {0,1,2,3,4}{
    \node[bubble] at (3.25+\i*0.52,5.45) {};
    \node[font=\tiny, text=black!55] at (3.25+\i*0.52,5.45) {\the\numexpr\i+1\relax};
  }
  \draw[redd!85!black, line width=1.0pt, densely dashed, rounded corners=2pt]
        (3.02,5.18) rectangle (5.72,7.58);
  \node[draw=redd!85!black, fill=redd!10, rounded corners=1.5pt, line width=0.5pt,
        font=\scriptsize\ttfamily, text=redd!75!black, inner sep=2pt]
        (bboxtag) at (4.55,4.78) {bbox};
  \draw[redd!70!black, line width=0.7pt, densely dotted] (5.72,5.55) -- (bboxtag.north);

  \node[numbadge] at (3.30,4.10) {30};
  \draw[gline] (3.70,4.12) -- (5.55,4.12);
  \draw[gline] (3.10,3.77) -- (5.55,3.77);
  \draw[gline] (3.10,3.50) -- (5.30,3.50);
  \draw[gline] (3.10,3.23) -- (5.55,3.23);
  \draw[gline] (3.10,2.96) -- (4.85,2.96);
  \foreach \i in {0,1,2,3,4}{
    \node[bubble] at (3.25+\i*0.52,2.50) {};
    \node[font=\tiny, text=black!55] at (3.25+\i*0.52,2.50) {\the\numexpr\i+1\relax};
  }

  \draw[redd!70!black, line width=0.9pt, -{Latex[length=2mm]}, densely dotted]
        (bboxtag.east) -- (6.35,4.78);

\end{tikzpicture}
\end{minipage}%
\hfill
\begin{minipage}[c]{0.58\linewidth}
\lstset{
  basicstyle=\scriptsize\ttfamily,
  columns=fullflexible,
  breaklines=true,
  keepspaces=true,
  showstringspaces=false,
  upquote=true,
  breakindent=0pt,
  breakautoindent=false,
  xleftmargin=2pt,
  literate={"}{{\textquotedbl}}1,
}
\begin{tcolorbox}[colback=black!2, colframe=black!30, boxrule=0.5pt, arc=1mm,
                  left=3pt, right=3pt, top=2pt, bottom=2pt, enhanced]
\begin{lstlisting}
{
  "meta": {
    "year": "2025", "subject": "Mathematics",
    "domain": "Geometry", "question_number": "28",
    "problem_type": "multiple-choice", "points": "4",
    "choices": ["$\sqrt{43}$", "$\sqrt{47}$",
                "$\sqrt{51}$", "$\sqrt{55}$",
                "$\sqrt{59}$"]
  },
  "image": {
    "multiple_choice": [
      {"image_name": "2025", "page_number": "19",
       "bbox": [863, 305, 1533,  577]},
      {"image_name": "2025", "page_number": "19",
       "bbox": [867, 565, 1511,  651]},
      {"image_name": "2025", "page_number": "19",
       "bbox": [875, 659, 1533, 1075]}
    ],
    "short_answer": [
      {"image_name": "2025", "page_number": "19",
       "bbox": [863, 305, 1533,  577]},
      {"image_name": "2025", "page_number": "19",
       "bbox": [875, 659, 1533, 1075]}
    ]
  },
  "answer": {
    "multiple_choice": "4",
    "short_answer": "$\sqrt{55}$"
  },
  "error": {
    "wrong_error_rate": "63.6",
    "choice_select_rate": ["9.0", "24.2", "10.6",
                           "36.4", "19.8"]
  }
}
\end{lstlisting}
\end{tcolorbox}
\end{minipage}
\caption{\textbf{Sample of structured metadata.} Each KCSAT-ML item links a schematic of its exam page (left; our own depiction, not the original exam) to structured annotation (right): attributes, page/bounding-box references, answers, and nationwide-cohort human-error statistics.}
\label{fig:metadata_example}
\end{figure*}

\section{Robustness and Validation of DRG}
\label{sec:drg_robustness}
Threshold sensitivity, comparison against a simpler metric, and a threshold-free check---all recomputed from the per-item logs of the main text, with no additional inference.

\paragraph{Conditions entering DRG.}
DRG contrasts each family's lowest-budget condition with its first TTS turn-on. That condition is the answer-only setting, except for five families (GPT-5, GPT-5-mini, GPT-5-nano, GPT-5.1, Gemini-3-Flash-Preview) exposing a vendor minimum reasoning tier below their first named budget level, for which we use that tier---the lowest budget offered as an ordinary decoding mode. The $22$ families are those with both conditions and a non-degenerate baseline; o4-mini, Gemini-2.5-Pro and Gemini-3-Pro-Preview are excluded, their low-budget runs being near-saturated or refusal-dominated (Sec.~\ref{sec:appendix_detail}).

\paragraph{Sensitivity to bin thresholds.}
The Easy and Hard bins are the tiers fixed \emph{a priori} in Section~\ref{sec:benchmark_construction} (Easy $\le 50\%$, Hard $\ge 76\%$ cohort error; $148$ and $78$ items), not tuned on results. Recomputing DRG over $16$ alternative cut-points (Table~\ref{tab:drg_threshold_sweep}; Hard thresholds are exclusive, so the reported tiers are the $50/75$ cell) leaves per-family rankings near-invariant ($\rho \ge 0.92$) and the HIGH/MID/LOW label intact for $77$--$100\%$ of families. Sign-stable families stay separated---Claude-Sonnet-4.5 is HIGH throughout ($[+19.4, +27.2]$), Qwen3-VL-2B negative ($[-14.8, -8.4]$)---and the same-accuracy Sonnet-4 vs.\ Sonnet-4.5 gap never falls below $13$\,pp (median $25$). DRG at the first turn-on also tracks DRG at maximum effort at $r = +0.943$, reproducing the $r{=}0.94$ of the main text.

\begin{table}[t]
\centering\small
\setlength{\tabcolsep}{4.5pt}
\begin{tabular}{lcccc}
\toprule
& \multicolumn{4}{c}{\textbf{Hard threshold (\%)}} \\
\cmidrule(lr){2-5}
\textbf{Easy (\%)} & $70$ & $75$ & $80$ & $85$ \\
\midrule
$40$ & .92\,/\,82 & .94\,/\,86 & .96\,/\,82 & .95\,/\,95 \\
$45$ & .96\,/\,91 & .98\,/\,95 & .97\,/\,77 & .96\,/\,86 \\
$50$ & .97\,/\,91 & \textbf{1.00\,/\,100} & .97\,/\,82 & .96\,/\,82 \\
$55$ & .95\,/\,86 & .98\,/\,86 & .96\,/\,82 & .96\,/\,77 \\
\bottomrule
\end{tabular}
\caption{\textbf{DRG under alternative bin thresholds.} Spearman $\rho$ of per-family DRG against the reported $50/75$ configuration (left) and the percentage of the $22$ families keeping their HIGH/MID/LOW label (right). The reported configuration (bold) is $1.00/100$ by construction.}
\label{tab:drg_threshold_sweep}
\end{table}

\paragraph{Comparison with a simpler metric.}
The natural alternative is the raw accuracy gap $\Delta\mathrm{Acc} = \Delta\mathrm{Acc}_{\text{Hard}} - \Delta\mathrm{Acc}_{\text{Easy}}$. It agrees with DRG only loosely ($r = +0.50$) and is dominated by floor effects: with wo.\,TTS accuracy in single digits the Easy-bin gain is large for almost every model, so $\Delta\mathrm{Acc}$ turns negative for $14$ of $22$ families (Claude-Sonnet-4.5: DRG $+24.3$, $\Delta\mathrm{Acc}$ $-0.5$) and the same cut-points yield $2/7/13$ HIGH/MID/LOW against DRG's $8/12/2$. The picture is unchanged under the uniform answer-only designation used by our released analysis code, which applies that condition to all $25$ families with a TTS pair: rank agreement $\rho = +0.85$, with $\Delta\mathrm{Acc}$ negative for $17$ of the $25$. SCF measures parity with the human cohort within a bin rather than raw accuracy, and is therefore invariant to that floor---which is why DRG is built on it.

\paragraph{Threshold-free convergence.}
\label{sec:drg_convergence}
The regression summary of Section~\ref{sec:three_regimes} describes the same behaviour with none of DRG's ingredients: a continuous OLS slope of model error on cohort error, no bins, no SCF, no thresholds. On the $17$ families whose low-budget condition is answer-only, its TTS-induced change correlates with DRG at $r = -0.81$ ($p<10^{-3}$)---families whose TTS clears Hard items are those whose error--difficulty slope flattens. It carries no signal for the five vendor-minimum-tier families ($r = -0.02$), which already emit reasoning tokens before turn-on so their slope barely moves, diluting the pooled figure to $r = -0.47$ ($p = 0.03$); under the uniform answer-only designation, which puts all $25$ families on a genuine reasoning-off baseline, the pooled correlation is $r = -0.79$ ($p < 10^{-5}$). Agreement at $|r| \approx 0.8$ wherever the contrast is a genuine reasoning-off/reasoning-on comparison indicates that DRG reflects a property of the model, not of the metric definition.

\section{Original-Format (Multiple-Choice) Evaluation}
\label{sec:mc_format}
The core set is evaluated in a short-answer format to remove multiple-choice guessing, whereas cohort error rates were measured on the original exam, which was partly multiple-choice. We check that this reformatting does not drive our conclusions from two directions: an untouched control subset, and a direct re-evaluation in the original format.

\paragraph{Untouched short-answer control.}
Of the $339$ core items, $136$ were originally short-answer (Easy/Medium/Hard $= 37/30/69$) and are therefore unaffected by any reformatting. The difficulty-conditioned pattern on this subset matches the full core set: pooled tier accuracy moves from $[49.9, 39.6, 25.7]$ to $[85.1, 78.1, 58.9]$ across wo./w.\,TTS, against $[45.7, 32.5, 25.3] \rightarrow [83.0, 69.7, 58.6]$ on all $339$ items---the same collapse--recovery shape, with the Hard tier recovering least in both. Per-family DRG computed on this subset correlates with DRG on the full core at Pearson $r = +0.72$ (Spearman $\rho = +0.67$). Under the uniform answer-only designation of our released analysis code (Appendix~\ref{sec:drg_robustness}), which places $25$ families on a common reasoning-off baseline, the same subset gives $[43.4, 34.0, 21.8] \rightarrow [89.0, 83.8, 70.3]$ and $r = +0.82$. Because this subset carries $69$ of the $78$ Hard items, the high-difficulty tail on which our claims rest is almost entirely un-reformatted.

\paragraph{Original multiple-choice re-evaluation.}
We additionally re-evaluate the $203$ originally-multiple-choice core items in their native format, options restored (Easy/Medium/Hard $= 111/83/9$), for Claude-Sonnet-4.5 and GPT-5.1 under both wo.\ and w.\,TTS ($812$ runs; extraction coverage $97.3\%$). In Table~\ref{tab:mc_format}, verdicts agree well above chance in every condition, DRG keeps its sign and magnitude, and the Easy${>}$Hard ordering holds wherever the model is not saturated. The reformatting's intended effect shows in the wo.\,TTS columns: with options present GPT-5.1 sits on the $20\%$ random-guess floor of five-option items, without them in single digits.

\paragraph{Scope of these DRG values.}
The re-evaluation pairs the two formats at an identical setting, taking each model's \emph{maximum} reasoning effort as the w.\,TTS condition rather than the first turn-on of Section~\ref{sec:drg}. The pair is therefore internally matched---what the format check requires---but not on the scale of the per-family DRG reported there, and the two should not be read against each other. The subset also skews Easy/Medium (only $9$ Hard items), making it complementary to the short-answer control above rather than a substitute.

\begin{table}[t]
\centering\small
\begin{adjustbox}{width=\linewidth}
\setlength{\tabcolsep}{4pt}
\begin{tabular}{lcccccccc}
\toprule
& \multicolumn{3}{c}{\textbf{wo.\,TTS}} & \multicolumn{3}{c}{\textbf{w.\,TTS}} & \multicolumn{2}{c}{\textbf{DRG}} \\
\cmidrule(lr){2-4}\cmidrule(lr){5-7}\cmidrule(lr){8-9}
\textbf{Model} & MC & SA & agr. & MC & SA & agr. & MC & SA \\
\midrule
Claude-Sonnet-4.5 & $47.2$ & $37.2$ & $68\%$ & $85.6$ & $72.5$ & $81\%$ & $+37.8$ & $+32.4$ \\
GPT-5.1 & $21.2$ & $9.9$ & $74\%$ & $99.5$ & $97.3$ & $98\%$ & $+13.3$ & $+12.6$ \\
\bottomrule
\end{tabular}
\end{adjustbox}
\caption{\textbf{Original multiple-choice format vs.\ our short-answer reformatting}, on the $203$ originally-multiple-choice core items. MC/SA: accuracy (\%) in each format; agr.: per-item verdict agreement. DRG here uses maximum reasoning effort and is not on the scale of Section~\ref{sec:drg}.}
\label{tab:mc_format}
\end{table}

\section{TTS Intervention Taxonomy and Same-Checkpoint Re-analysis}
\label{sec:tts_taxonomy}
The wo./w.\,TTS contrast is not a single intervention across families. We categorise each family by what turning TTS on actually changes (Table~\ref{tab:tts_taxonomy}), then re-derive the main causal claims on the subset where the checkpoint is held fixed. Types (i), (ii) and (iv) hold the checkpoint fixed; only (iii) replaces it. Of the $34$ families reported in this paper, $26$ are checkpoint-fixed; of the $22$ carrying DRG values, $14$ are.

\begin{table}[t]
\centering\small
\begin{tabularx}{\linewidth}{@{}>{\raggedright\arraybackslash}X@{}}
\toprule
\textbf{(i) Budget tier} ($7$) --- same checkpoint at a higher vendor reasoning-effort tier \\
\addlinespace[1pt]
GPT-5, -Mini, -Nano, GPT-5.1, o4-Mini, Gemini-3-Flash, -Pro \\
\midrule
\textbf{(ii) Mode toggle} ($10$) --- same checkpoint with a vendor reasoning mode switched on \\
\addlinespace[1pt]
Claude-Sonnet-4, -4.5, -Haiku-4.5, -Opus-4.5, Gemini-2.5-Flash, -Flash-Lite, -Pro, HCX-Seed-14B, -Think-32B, -Omni-8B \\
\midrule
\textbf{(iii) Checkpoint swap} ($8$) --- a separately trained Thinking checkpoint replaces the Instruct one \\
\addlinespace[1pt]
Qwen3-4B, -30B-A3B, Qwen3-VL-2B, -4B, -8B, -30B-A3B, -32B, Kanana-2-30B-A3B \\
\midrule
\textbf{(iv) Prompt control} ($9$) --- same checkpoint, answer-only vs.\ solve prompt \\
\addlinespace[1pt]
GPT-OSS-20B, -120B, Exaone-4.0-32B, Solar-10.7B, -Pro, Midm-2.0, SKT-AX-4.0, -Light, HCX-Vision-3B \\
\bottomrule
\end{tabularx}
\caption{\textbf{What turning TTS on changes, per family.} Family counts in parentheses; the models affected are listed under each type. Only (iii) changes the weights.}
\label{tab:tts_taxonomy}
\end{table}

\paragraph{Anti-scaling holds within a fixed checkpoint.}
The anti-scaling result of Section~\ref{sec:two_faces} contrasts Qwen3-VL-32B-Instruct with Qwen3-VL-8B-Thinking, a type-(iii) pair. Both halves of the claim survive without any checkpoint swap. On the $23$ items with cohort error $\ge 93\%$, Qwen3-VL-32B-Instruct solves $0/23$ under answer-only decoding but $9/23$ when the \emph{same} Instruct checkpoint is given a solve prompt---matching the $9/23$ of the reasoning-optimised 8B-Thinking checkpoint, so the recovery does not require reasoning-optimised weights. And the size inversion reappears within the Instruct family alone: 8B-Instruct with a solve prompt reaches $6/23$, against $0/23$ for the $4\times$ larger 32B-Instruct without one.

\paragraph{Overthinking holds within a fixed checkpoint.}
Restricting the overthinking count of Section~\ref{sec:two_faces} to checkpoint-fixed interventions, the wo.\,TTS-correct$\to$w.\,TTS-incorrect direction persists across four architecturally distinct families: GPT-OSS-120B ($4$ items, already type (iv) in the main text), Claude-Sonnet-4 ($2$, mode toggle), Qwen3-VL-32B ($3$, prompt-level) and Qwen3-30B-A3B ($3$, prompt-level).

\paragraph{DRG classification under checkpoint swaps.}
For the eight type-(iii) families we recompute DRG \emph{within} the Instruct checkpoint (off $=$ Instruct answer-only, on $=$ the same checkpoint with a solve prompt). Five of the eight keep their HIGH/MID/LOW label, and the three that move go in \emph{both} directions---Qwen3-VL-4B $-8.7\!\to\!+0.9$ (LOW$\to$MID, a boundary case), Qwen3-VL-30B-A3B $+9.4\!\to\!+15.8$ (MID$\to$HIGH) and Kanana-2 $+0.8\!\to\!-16.6$ (MID$\to$LOW)---so checkpoint swaps add noise rather than a systematic bias. The clearest LOW case, Qwen3-VL-2B, stays LOW within the fixed checkpoint ($-9.7\!\to\!-7.5$). We accordingly scope strictly causal claims to checkpoint-fixed families and treat the cross-family pattern as external validity.

\section{LLM-Judge Validation}
\label{sec:judge_validation}
Answers are graded by an LLM equivalence judge (\texttt{GPT-5-2025-08-07}, Sec.~\ref{sec:benchmark_construction}; prompt in Figure~\ref{fig:prompt}) because model outputs vary in surface form---``$35$'', ``$\boxed{35}$'', ``the answer is $35$''---which string matching cannot reconcile, while hand-grading every run is infeasible; supplying the gold answer and asking only for a \texttt{Correct}/\texttt{Incorrect} verdict is standard in recent evaluation work~\citep{chen2025xverify, liu-etal-2025-compassverifier}. We validate the judge against a deterministic oracle on the subset where one exists.

\paragraph{Agreement with rule-based grading.}
Of the $339$ core items, $227$ ($67\%$) have integer ground-truth answers, for which exact matching after normalisation requires no human labor. Over the $95{,}072$ runs on these items, deterministic rules recover a final answer for $70{,}431$ ($74.1\%$ coverage); on those the rule-based verdict agrees with the judge on $\mathbf{99.31\%}$ of runs. The $483$ disagreements split into $351$ judge-lenient ($0.50\%$) and $132$ judge-strict ($0.19\%$), so the judge shows no systematic tendency to accept or to reject.

\paragraph{The residual disagreement is extraction noise.}
Table~\ref{tab:judge} breaks the comparison down by extraction rule. The two unambiguous rules---a \verb|\boxed{}| capture, or a response consisting of the bare answer---disagree with the judge on $20$ of $44{,}982$ runs ($99.96\%$), while the two heuristic rules, which read a number out of an answer-like phrase or off the final line, account for $463$ of the $483$ disagreements ($96\%$). The typical case is a heuristic picking up an intermediate quantity from a long derivation where the judge reads the stated final answer, which places the residual error on the deterministic side rather than the judge's.

\paragraph{Agreement does not track the condition under test.}
Split by condition (Table~\ref{tab:judge}), agreement drops by half a point when reasoning is on. The gap is a composition effect: reasoning-on responses are longer, so the heuristic rules fire on $46\%$ of them against $26\%$ with reasoning off. Restricted to unambiguous extractions the two conditions are indistinguishable ($99.958\%$ vs.\ $99.953\%$). Judge disagreement of under $1$\,pp, evenly distributed across conditions, cannot manufacture the wo./w.\,TTS contrasts of tens of points that DRG is built on. The oracle covers integer answers only; the remaining $112$ core items, whose answers are fractions or symbolic expressions, are precisely the cases that motivate an equivalence judge in the first place.

\begin{table}[t]
\centering\small
\begin{tabular}{@{}lrr@{}}
\toprule
& $n$ & \textbf{Agreement} \\
\midrule
\multicolumn{3}{@{}l}{\emph{By extraction rule}} \\
\quad Bare answer & $10{,}483$ & $99.98\%$ \\
\quad \verb|\boxed{}| capture & $34{,}499$ & $99.95\%$ \\
\quad Answer phrase & $13{,}343$ & $98.89\%$ \\
\quad Final line & $12{,}106$ & $97.40\%$ \\
\addlinespace[3pt]
\multicolumn{3}{@{}l}{\emph{By condition}} \\
\quad Reasoning off & $35{,}201$ & $99.58\%$ \\
\quad Reasoning on & $35{,}230$ & $99.05\%$ \\
\midrule
Overall & $70{,}431$ & $99.31\%$ \\
\bottomrule
\end{tabular}
\caption{\textbf{LLM judge vs.\ deterministic grading} on the $70{,}431$ runs over integer-answer core items where a rule-based answer can be extracted. Disagreement concentrates in the two heuristic extraction rules, not in any condition.}
\label{tab:judge}
\end{table}

\section{Alternative Difficulty Proxies}
\label{sec:proxies}
Section~\ref{sec:diff_signal} weighs the cohort error rate against examiner points. We extend that comparison to every model-independent item signal available to us. The dependent variable is an item's mean accuracy across configurations, taken separately for reasoning-off and reasoning-on runs; the predictors are the cohort error rate, the examiner point value, the item's exam track or elective subject (an eight-level categorical covering the pre-2022 Ga/Na and A/B tracks and the post-reform Common, Probability \& Statistics, Calculus and Geometry), the length of the OCR problem statement ($\log_{10}$ characters), and whether the item carries a figure ($84$ of $339$, labelled by hand). Solution length, a natural further candidate, cannot be computed---no official step-by-step solutions are released---so the model-independent statement length stands in for it. This regression pools conditions differently from Section~\ref{sec:diff_signal}, so its absolute $R^2$ is not directly comparable with the values reported there; what matters here is each signal's relative contribution.

\paragraph{The cohort error rate carries what item metadata cannot.}
Removing one signal at a time from the five-signal model (Table~\ref{tab:proxies}) isolates the part of the fit that only that signal supplies. The cohort error rate accounts for the largest share under both conditions, and it is the only signal whose unique contribution \emph{grows} when TTS is turned on. Examiner points and figure presence are almost entirely recoverable from the remaining signals, so a benchmark carrying only such metadata would retain little of the difficulty information the cohort statistic provides. The one signal that approaches it under TTS is statement length, a property of how an item is written rather than of how examinees performed on it; even with statement length in the model, the cohort error rate's unique contribution remains the largest.

\paragraph{Model-derived signals predict better but cannot anchor the axis.}
Two signals computed from model outputs do predict per-item accuracy better than any of the five: output-token length ($R^2 = .500/.556$ for wo./w.\,TTS) and ensemble disagreement, the spread of correctness across evaluated families ($.793/.774$). Both are functions of the outputs being predicted. Ensemble disagreement is structurally circular---for mean correctness $p$ it tracks $p(1-p)$---and its $R^2$ falls to $.47$--$.60$ once the disagreement is taken from the other decoding condition. Adopting ``what models disagree on'' as the difficulty axis would define difficulty in terms of model weakness and leave our central question---whether a model fails where humans fail---undefinable. The cohort error rate is the only signal external to every evaluated model, which is what lets it anchor SCF, DRG, and the below-cohort regime.

\begin{table}[t]
\centering\small
\setlength{\tabcolsep}{4pt}
\begin{tabular}{@{}lrrrr@{}}
\toprule
& \multicolumn{2}{c}{\textbf{wo.\,TTS}} & \multicolumn{2}{c}{\textbf{w.\,TTS}} \\
\cmidrule(lr){2-3}\cmidrule(lr){4-5}
\textbf{Signal removed} & $R^2$ & $\Delta$ & $R^2$ & $\Delta$ \\
\midrule
None (all five) & $.436$ & --- & $.415$ & --- \\
\addlinespace[2pt]
Cohort error rate & $.354$ & $\mathbf{-.081}$ & $.307$ & $\mathbf{-.108}$ \\
Track / subject & $.382$ & $-.054$ & $.368$ & $-.047$ \\
Statement length & $.398$ & $-.038$ & $.325$ & $-.091$ \\
Examiner points & $.411$ & $-.025$ & $.413$ & $-.003$ \\
Figure presence & $.430$ & $-.006$ & $.403$ & $-.012$ \\
\bottomrule
\end{tabular}
\caption{\textbf{Leave-one-out ablation over the five model-independent signals.} $R^2$ for per-item mean accuracy after dropping each signal from the full model; $\Delta$ is the resulting loss, i.e.\ the part of the fit that only that signal supplies. Larger $|\Delta|$ means less recoverable from the other signals.}
\label{tab:proxies}
\end{table}

\end{document}